\documentclass[preprint,12pt,authoryear]{elsarticle}

\usepackage{amssymb}
\usepackage{fullpage}

\usepackage{amsmath}
\usepackage{booktabs}
\usepackage{url}
\usepackage{subcaption}
\usepackage{multirow}
\usepackage[ruled,vlined]{algorithm2e}
\usepackage{xfrac}
\usepackage{xcolor}

\usepackage{blindtext}
\usepackage{hyperref}

\usepackage{tikz}
\usetikzlibrary{positioning}
\usetikzlibrary{arrows.meta}
\usetikzlibrary{plotmarks}
\definecolor{lightgray}{rgb}{0.83, 0.83, 0.83}

\newcommand{\bm}[1]{%
    \boldsymbol{#1}
}

\newcommand{\new}[1]{#1}

\journal{Applied Soft Computing}

\begin{document}

\begin{frontmatter}



\title{Graph Neural Networks for the Offline Nanosatellite Task Scheduling Problem}


\author[1]{Bruno Machado Pacheco}
\ead{bruno.m.pacheco@posgrad.ufsc.br}

\author[1]{Laio Oriel Seman}
\ead{laio@ieee.org}

\author[2]{Cezar Antônio Rigo}
\ead{cezar.a.rigo@gmail.com}

\author[1]{Eduardo Camponogara}
\ead{eduardo.camponogara@ufsc.br}

\author[2]{Eduardo Augusto Bezerra}
\ead{eduardo.bezerra@ufsc.br}

\author[3]{Leandro dos Santos Coelho}
\ead{leandro.coelho@ufpr.br}

\affiliation[1]{organization={Department of Automation and Systems Engineering, Federal University of Santa Catarina (UFSC)},
            city={Florianópolis},
            state={SC},
            country={Brazil}}

\affiliation[2]{organization={Department of Electrical Engineering, Federal University of Santa Catarina (UFSC)},
            city={Florianópolis},
            state={SC},
            country={Brazil}}

\affiliation[3]{organization={Department of Electrical Engineering, Federal University of Parana (UFPR)},
            city={Curitiba},
            state={PR},
            country={Brazil}}

\begin{abstract}
This study investigates how to schedule nanosatellite tasks more efficiently using Graph Neural Networks (GNNs).
In the Offline Nanosatellite Task Scheduling (ONTS) problem, the goal is to find the optimal schedule for tasks to be carried out in orbit while taking into account Quality-of-Service (QoS) considerations such as priority, minimum and maximum activation events, execution time-frames, periods, and execution windows, as well as constraints on the satellite's power resources and the complexity of energy harvesting and management.
This study explores the use of Graph Neural Networks (GNNs) as primal heuristics for the ONTS problem, as this class of deep learning model has been effectively applied to optimization problems such as the traveling salesman, scheduling, and facility placement.
We investigate whether GNNs can learn the complex structure of the ONTS problem with respect to feasibility and optimality of candidate solutions.
Furthermore, we evaluate using GNN-based heuristic solutions to provide better solutions (w.r.t. the objective value) to the ONTS problem and reduce the optimization cost.
Our experiments show that GNNs are not only able to learn feasibility and optimality for instances of the ONTS problem, but they can generalize to harder instances than those seen during training, addressing the data acquisition cost of training deep learning models on optimization problems.
On top of that, the GNN-based heuristics improved the expected objective value of the best solution found under the time limit in 45\%, and reduced the expected time to find a feasible solution in 35\%, when compared to the SCIP (Solving Constraint Integer Programs) solver in its off-the-shelf configuration.
\end{abstract}



\begin{keyword}
Scheduling \sep Graph Neural Network \sep Combinatorial Optimization \sep  Nanosatellite \sep Deep Learning



\end{keyword}

\end{frontmatter}


\section{Introduction} \label{sec:intro}

Nanosatellites have attracted attention from industry and academia for over a decade, primarily due to their reduced development and launch costs~\citep{shiroma_cubesats_2011,lucia_computational_2021,Nagel2020,saeed_cubesat_2020}. However, these miniaturized spacecraft face significant operational challenges due to their inherent resource constraints, particularly in task scheduling and resource management.

The Offline Nanosatellite Task Scheduling (ONTS) problem addresses these challenges by optimizing the allocation of satellite resources during the mission planning phase. This optimization is particularly complex due to the diverse operational characteristics of satellite payloads, including varying power consumption profiles, data handling requirements, and specific operational windows. Furthermore, the reliance on solar power as the primary energy source introduces considerable variability in energy supply, affected by factors such as cyclical eclipses in Low Earth Orbit (LEO), varying solar incidence angles, spacecraft attitude, and size.

As an offline planning tool, ONTS assists engineers in simulating and evaluating performance before launch. The problem must address five fundamental scheduling requirements: \textit{i}) non-preemptive tasks that must be completed without interruption once initiated; \textit{ii}) task periodicity within specific thresholds; \textit{iii}) minimum and maximum task startups within an orbit to meet Quality-of-Service (QoS) requirements; \textit{iv}) task execution windows for precise operations; and \textit{v}) battery state-of-charge management balancing energy consumption with generation.

Traditional approaches to solving the ONTS problem have employed various mathematical optimization techniques, progressing from Integer Programming (IP)~\citep{rigo_task_2021} to Mixed-Integer Linear Programming (MILP)~\citep{rigo_nanosatellite_2021,seman_energy-aware_2022} and Continuous-Time formulations~\citep{camponogara_continuous-time_2022}. However, due to the (Non-deterministic Polynomial time)-hard nature of the problem, efficiently solving large instances—involving many tasks or extended planning horizons—remains an open research challenge.

Meanwhile, several recent investigations~\citep{bengio_machine_2021,karimi-mamaghan_machine_2022,pacheco2023deeplearningbased,han_gnn-guided_2023} have considered machine learning tools to address combinatorial optimization problems, such as the single machine problem~\citep{parmentier_structured_2023}, resource-constrained project scheduling~\citep{guo_prediction_2023}, and knapsack problems~\citep{yang_learning_2022}.
Graph Neural Networks (GNNs), in particular, have gained popularity in recent years to solve combinatorial optimization problems when the underlying structure can be represented as a graph~\citep{zhang_survey_2023}, which is the case for MILP~\citep{khalil_mip-gnn_2022}.

This paper investigates the novel application of GNNs for the ONTS problem.
More specifically, we consider two research questions:
\begin{itemize}
    \item Can a graph neural network learn the structure of the ONTS problem?
    \item Is a GNN-based heuristic effective (fast and accurate) for the ONTS problem?
\end{itemize}
To address these two questions, we propose two sets of experiments.
First, we train GNNs to predict the feasibility and optimality of candidate solutions given an instance of the problem.
In the second set of experiments, we train GNNs to generate candidate solutions given problem instances and use the resulting models to build matheuristics~\citep{boschetti_matheuristics_2022}.
Our results indicate that the proposed GNN can learn the feasibility and optimality of candidate solutions, even when given instances that are larger (and harder) than those seen during training.
Furthermore, GNN-based matheuristics effectively provide high-quality solutions for large problems, overcoming the MILP solver (SCIP) both in the time to generate feasible solutions and the quality of the solutions found within limited time.

Everything considered, this paper takes a fundamentally different approach compared to previous works. Rather than proposing a new standalone algorithm, our work aims to develop a supportive heuristic component that can enhance existing solution methods. This is accomplished through three main aspects: (\textit{i}) the GNN-based heuristic is designed specifically to improve and accelerate the performance of traditional exact methods or sophisticated ones like branch-and-price, not to replace them; (\textit{ii}) it serves as an integrated component that works in conjunction with established algorithms, providing high-quality initial solutions to speed up convergence; and (\textit{iii}) it maintains the mathematical guarantees of the host algorithm while improving its computational efficiency.
In summary, the main contributions of this paper are as follows:
\begin{itemize}
\item The introduction of SatGNN, a pioneering GNN architecture suitable to the ONTS problem;
\item A showcase of SatGNN's impressive performance in accurately classifying feasibility, predicting optimality, and generating effective heuristic solutions;
\item A demonstration of GNN's robust generalization abilities for optimization problems, particularly when dealing with larger ONTS instances;
\item An implication of the potential for a transformative impact on complex space mission scheduling.
\end{itemize}


The remainder of this paper is organized as follows:
Section \ref{sec:rel-work} provides an overview of the related literature.
Section \ref{sec:onts} describes the problem in detail, providing context and background information and formulating the optimization problem.
Section \ref{sec:gnns} provides the necessary theoretical background on GNNs and their application to linear optimization problems.
The data acquisition, the proposed GNN architecture, and the heuristics for optimization problems are presented in Section \ref{sec:meth}.
The experiments that tackle the two research questions and the results are presented in Section \ref{sec:exps}.
Finally, Sections \ref{sec:discussion} and \ref{sec:conclusion} discuss the key findings and provide concluding remarks and future research directions.

\section{Related Work}\label{sec:rel-work}

The ONTS problem has received significant contributions in recent years, with different approaches and formulations.
At the same time, machine learning applications for optimization problems have become a prolific area of research.
This section briefly overviews our paper's most influential works, which are also summarized in Table~\ref{tab:related-work}.

\begin{table}[htbp]
    \centering
    \caption{Summary of related work}
    \label{tab:related-work}
    \footnotesize
    \begin{tabular}{p{2.5cm}p{6cm}p{4cm}p{2.5cm}}
        \toprule
        \textbf{Reference} & \textbf{Main Contribution} & \textbf{Methodology} & \textbf{Application} \\
        \midrule
        \multicolumn{4}{l}{\textit{Nanosatellite Task Scheduling}} \\
        \midrule
        \citet{wang_dynamic_2013} & Dynamic scheduling of emergency tasks & Mathematical programming with multiple objectives & Emergency tasks \\
        \citet{wang_pure_2016} & Handling cloud blockage uncertainties & Approximate integer linear programming & Visibility constraints \\
        \citet{NIU2018813} & Multi-satellite scheduling under emergency conditions & Integer programming + genetic algorithm & Emergency response \\
        \citet{rouzot:hal-04430171} & Scheduling under tight nanosatellite constraints & Constraint programming & Nanosatellite missions \\
        \citet{rigo_task_2021} & Power-aware task allocation & Integer programming & Power optimization \\
        \citet{seman_energy-aware_2022} & Constellation task scheduling & Mixed-integer linear programming & Nanosatellite constellations \\
        \citet{rigo_branch-and-price_2022} & Column generation approach & Branch-and-price & Task scheduling \\
        \midrule
        \multicolumn{4}{l}{\textit{Machine Learning for Scheduling}} \\
        \midrule
        \citet{hopfield_neural_1985} & First ML application to optimization & Hopfield neural networks & TSP \\
        \citet{wang_complex_2021} & Heuristic scheduling solutions & Reinforcement learning (MDP) & General scheduling \\
        \citet{FENG2024102362} & Static constraint satisfaction & Deep learning + optimization & Satellite scheduling \\
        \midrule
        \multicolumn{4}{l}{\textit{GNNs for Combinatorial Optimization}} \\
        \midrule
        \citet{khalil_learning_2017} & Graph-based CO solutions & GNN-based greedy heuristic & Graph problems \\
        \citet{gasse_exact_2019} & MILP problem embedding & Bipartite graph representation & General MILP \\
        \citet{ding_accelerating_2020} & Variable selection in B\&B & GNN-guided branching & MILP solving \\
        \citet{nair_neural_2020} & Neural Diving approach & GNN-based partial solutions & MILP solving \\
        \citet{han_gnn-guided_2023} & Trust region optimization & GNN-guided solution space exploration & MILP solving \\
        \bottomrule
    \end{tabular}
\end{table}

\subsubsection*{Nanosatellite task scheduling}

Mathematical programming is consolidated as a practical approach for satellite scheduling problems.
\citet{wang_dynamic_2013} formulate the problem of dynamic scheduling emergency tasks as a mathematical program with multiple objectives.
\citet{wang_pure_2016} handle the uncertainties from cloud blockage with respect to the satellite visibility and propose an approximate integer linear program to schedule tasks, which is solved by the authors using a branch and cut approach.
Considering multiple satellites, \citet{NIU2018813} propose an integer program model and a genetic algorithm to provide feasible solutions faster under emergency conditions.

Targeting small satellites, \citet{rouzot:hal-04430171} formulated the problem using constraint programming to generate schedules that satisfy the multiple tight requirements that a nanosatellite mission imposes.
\citet{rigo_task_2021} formulated the problem of maximizing task allocation under power availability constraints as an integer programming (IP) problem, enabling the generation of optimal schedules.
\citet{seman_energy-aware_2022} extended the approach by \citet{rigo_task_2021} and proposed a Mixed-Integer Linear Programming (MILP) formulation to tackle the task scheduling in constellations of nanosatellites.
\citet{camponogara_continuous-time_2022} applied the Multi-Operation Sequencing with Synchronized Start Times (MOS-SST) representation to the problem, maintaining an MILP formulation but extending the problem to continuous time.
The evolution of formulations and methodologies highlights the continuous efforts to find the most efficient and effective solutions to the ONTS problem.

More recently, given the difficulty in solving complex instances of the ONTS problem, \citet{rigo_branch-and-price_2022} proposed a Dantzig-Wolfe decomposition, coupled with a branch-and-price (B\&P) methodology, which was further adapted in a heuristic solution through the use of a genetic algorithm~\citep{SEMAN2023110475}.
This approach was designed to build a unique column-based formulation that produces feasible and optimal schedules.
In addition, they leveraged the Dynamic Programming (DP) technique to identify optimal columns and speed up the whole process.
Their computational experiments significantly improved overall solution time compared to a commercial MILP solver, with a 70\% reduction in computation time.
Despite these promising results, the B\&P method introduces specific challenges and limitations that preclude its universal application.
One significant drawback is its heavy reliance on the suitability of the decomposition, which is notoriously problem-specific.
Currently, no known method allows practitioners to \emph{a priori} ascertain the presence of a decomposable structure in a given problem~\citep{kruber_learning_2017}.
This lack of a deterministic approach to identifying suitable problems adds a layer of complexity to the application of B\&P.
Additionally, the complexity inherent in the decomposition and the optimization algorithm can result in a high development undertaking.
The intricate nature of these processes requires specialized knowledge and expertise, potentially rendering the B\&P method prohibitive for particular applications or settings.
Consequently, while the B\&P methodology offers substantial improvements in computational efficiency for specific problem instances, its applicability may be limited by these challenges, emphasizing the need for caution in considering it as a universal solution for all optimization problems.

The literature demonstrates a clear progression in solution methodologies for the ONTS problem, from classical mathematical programming to sophisticated decomposition approaches. While the B\&P methodology has demonstrated substantial computational advantages, its effectiveness remains contingent upon problem-specific characteristics and implementation expertise. These limitations underscore the necessity for developing more generalizable solution approaches that maintain computational efficiency while reducing implementation complexity.

\subsubsection*{Machine learning for offline task scheduling}

Machine learning applications for optimization problems were first proposed in 1985, applying Hopfield neural networks to the Travelling Salesperson Problem (TSP)~\citep{hopfield_neural_1985}.
Due to the hardware limitations of the time~\citep{smith_neural_1999}, neural network applications to optimization problems have stayed out of the spotlight up to recent years, when developments in novel architectures, more powerful hardware, and promising published results have attracted the attention of the research community back to this area~\citep{bengio_machine_2021}.

For scheduling problems, many authors have proposed reinforcement learning strategies to make the scheduling decisions, even in offline settings.
A series of works have focused on training models to provide completely heuristic solutions, modeling the scheduling problem as a markov decision process~\citep{wang_complex_2021,tassel2021reinforcement,NEURIPS2020_11958dfe}.
Another research direction with significant contributions has been on modeling uncertainty through reinforcement learning and modifying the parameters of the scheduling IP formulation based on the model's output~\citep{Kenworthy_Nayak_Chin_Balakrishnan_2022,gomes_reinforcement_2017}.
To the best of our knowledge, no works in scheduling have proposed learning-based matheuristics, for example, as is done for vertex coloring and set covering in \citet{kruber_learning_2017}.

This is particularly true for satellite task scheduling problems, where many works focused on reinforcement learning policies for generating a heuristic solution for the schedule~\citep{10004750,rs13122377,9998480,9152114}.
To the best of our knowledge, the work by~\citet{FENG2024102362} is the only one to use a deep learning model interacting directly with the solver of a satellite task scheduling problem.
The authors train a deep learning model to solve the problem's static constraints, alleviating the optimizer's job, which aims to satisfy the dynamic constraints of the problem.

The applications of machine learning in scheduling problems have predominantly focused on reinforcement learning strategies, both for direct heuristic solutions and uncertainty modeling. However, the integration of learning-based matheuristics with satellite task scheduling remains largely unexplored, particularly in contrast to other combinatorial optimization domains. This gap in the literature presents opportunities for novel methodological contributions.

\subsubsection*{GNNs for combinatorial optimization}

One of the drivers of the recent interest in machine learning for Combinatorial Optimization (CO) problems was the proposition of structured artificial neural networks able to handle the symmetries that optimization problems have~\citep{cappart_combinatorial_2022}, such as the invariance to permutation of constraints.
A significant example of such property for CO is the work of \citet{khalil_learning_2017}, which developed a greedy heuristic for CO problems on graphs using a GNN.
\citet{gasse_exact_2019} proposed to embed linear programming problems as a bipartite graph, generalizing the use of GNNs to any MILP.
The results presented in these works indicate that GNNs can learn the structures of MILP problems.

Many authors proposed using GNNs to predict solution values, aiming at using the learned model in heuristic solutions.
\citet{ding_accelerating_2020} proposed using such GNNs to guide the Branch and Bound (B\&B) algorithm towards the variables the model had the most difficulty learning.
\citet{khalil_mip-gnn_2022} use the GNNs trained to predict biases of the binary variables to perform node selection and warm starting.
\citet{nair_neural_2020}, in their approach named \emph{Neural Diving}, proposed to train the model to provide partial solutions, which generate sub-problems through variable fixing.
The multiple sub-problems are then solved using off-the-shelf solvers.
\citet{han_gnn-guided_2023} compare the fixing approach of \citet{nair_neural_2020} to their proposed approach of defining a trust region around the partial solution generated by the GNN, showing significant improvements over off-the-shelf solvers like SCIP and Gurobi.
These works highlight how promising building GNN-based heuristic approaches to MILP can be.

The literature evidences the efficacy of GNN architectures in addressing combinatorial optimization problems, particularly through their integration with traditional exact methods such as B\&B. Recent developments in solution prediction and variable fixing strategies have yielded promising computational results. Nevertheless, the application of these techniques remains an emerging field, with significant potential for theoretical and practical advances.

\subsubsection*{}


To the best of our knowledge, while metaheuristic approaches like genetic algorithms have been applied to nanosatellite task scheduling~\citep{SEMAN2023110475}, our work is the first to propose a matheuristic specifically designed to enhance exact solution methods. Unlike metaheuristics that operate independently but cannot guarantee global optimality, our approach is engineered to provide high-quality initial solutions that accelerate the convergence of exact methods while maintaining their optimality guarantees. 
Furthermore, machine learning applications to MILP has only recently attracted intense attention, as confirmed by the novelty of the works presented above and the lack of learning-based components in open-source solvers such as SCIP and CPLEX. In other words, although promising, applying GNNs to MILP still needs to be well-understood and is not guaranteed to yield positive results. In this context, the present work also contributes to the area of machine learning and combinatorial optimization research by validating novel techniques that complement rather than replace exact methods.

\section{Offline Nanosatellite Task Scheduling (ONTS)}\label{sec:onts}

\begin{figure}[h]
    \centering
    \includegraphics[width=0.5\textwidth]{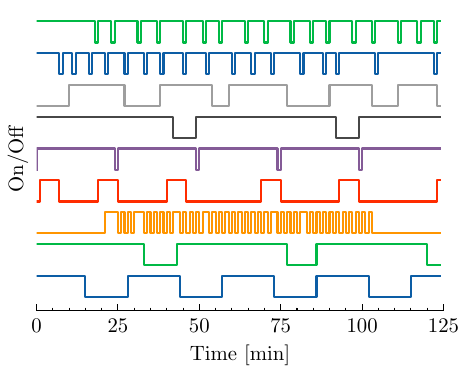}
    \caption{Illustration of an optimum scheduling for 9 tasks on a horizon of 125 time steps. Each color represents the execution of the different tasks.}
    \label{fig:example-scheduling}
\end{figure}

\begin{figure}[h]
    \centering
    \includegraphics[width=0.4\textwidth]{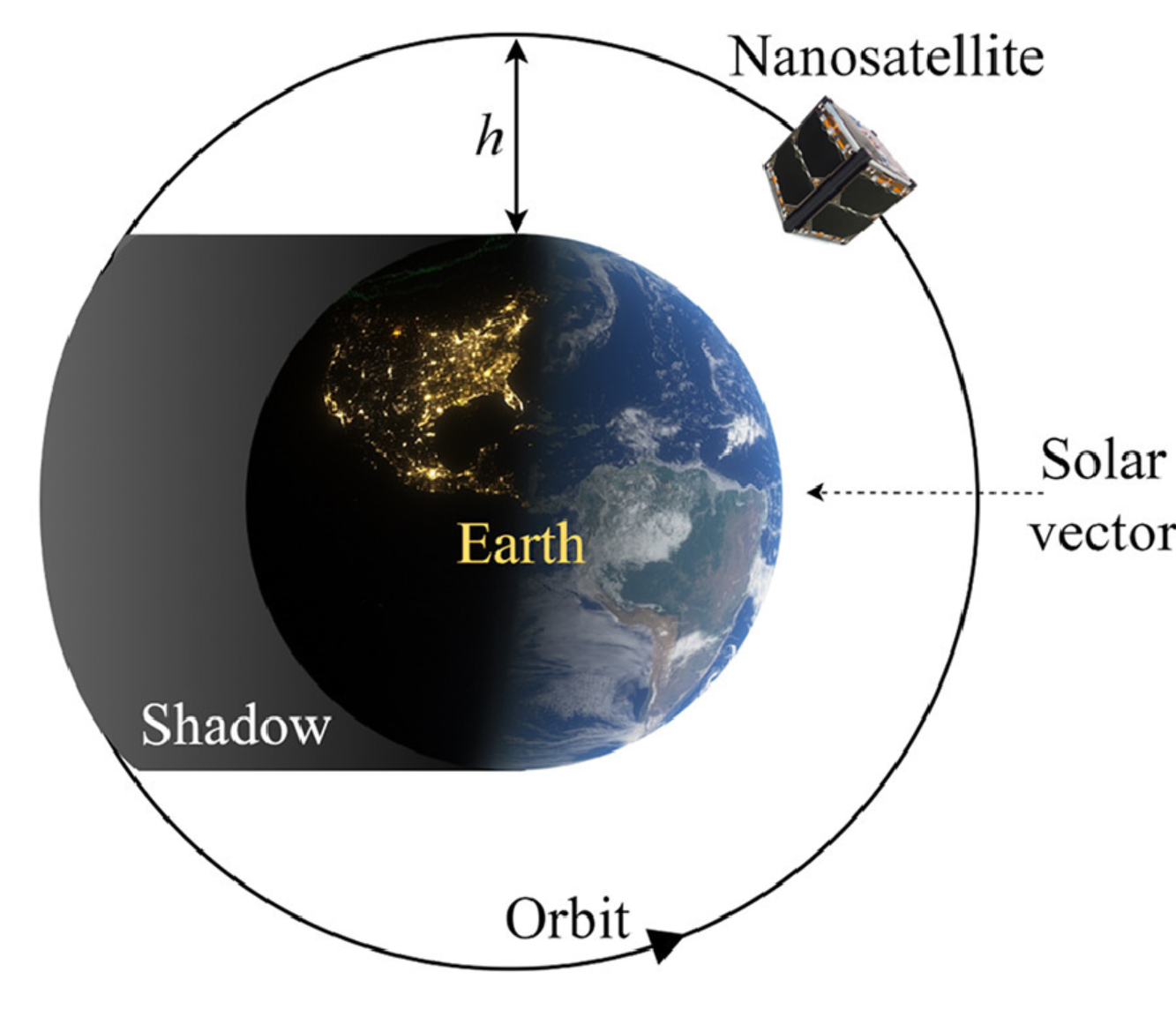}
    \caption{Illustration of a nanosatellite's orbit around Earth. Image from \citet{rigo_branch-and-price_2022}.}
    \label{fig:onts-orbit}
\end{figure}

Nanosatellite scheduling problems concern the decisions on each task's start and finish time.
The tasks usually require periodic execution and during restricted moments along the orbit.
Besides time, energy availability through orbit is a crucial resource to consider.
Figure \ref{fig:example-scheduling} shows an example of optimal scheduling, in which each job is represented by a different color and the activation and deactivations are shown through the steps of the signals.
Proper scheduling must account for energy management, so the tasks do not draw more energy than the system can provide and the battery is not depleted before the end of the mission.
Energy management is a difficult task since the nanosatellite draws power from its solar panels, with the energy availability depending on the attitude of the nanosatellite (which affects the orientation of the solar panels) and the trajectory with respect to Earth's shadow, as illustrated in Figure \ref{fig:onts-orbit}.

\subsection{MILP Formulation}\label{sec:problem}

A formulation that considers the realistic constraints and objectives of ONTS was proposed by \citet{rigo_task_2021} and is presented here.
For reference, all symbols (sets, variables, and parameters) used in our formulation are summarized in \ref{appx:tab-symbols}.
Given a set of jobs $\mathcal{J}=\{0,...,J\}$ that represents the tasks, and a set of time units $\mathcal{T}=\{0,...,T\}$ that represents the scheduling horizon, variables \[
\bm{x} = \left(x_{j,t}\right)_{\substack{j=1,\ldots,J \\ t = 1,\ldots,T}}
\] represent the allocation of jobs $j\in \mathcal{J}$ at times $t\in\mathcal{T}$, i.e., $x_{j,t}=1$ indicates that job $j$ is scheduled to execute at time $t$.
Naturally, $x_{j,t}$ are binary variables, i.e., $\bm{x} \in \{0,1\}^{JT}$.

It is assumed that there exists a priority for every job, which is defined by $\bm{u} = (u_1,\ldots,u_{J})$.
The goal is to maximize the mission's Quality of Service (QoS) \eqref{eq:qos}, which is the sum of the allocations weighted by the priorities,
\begin{align}\label{eq:qos}
{\rm QoS}(\bm{x};\bm{u}) =
	\sum_{j=1}^{J} \sum_{t=1}^{T} {u}_{j} x_{j,t}
\end{align}

To formalize the ONTS problem of maximizing QoS, we follow the formulation proposed by \citet{rigo_task_2021}.
The following constraints are added to ensure that the scheduling respects the requirements and specificities of each job:
\begin{subequations}\label{eq:qos-constraints}
    \begin{align}
        &\phi_{j,t} \geq x_{j,t}, &~\forall j\in\mathcal{J},\, t = 1 \label{phiA} \\
        &\phi_{j,t} \geq x_{j,t} - x_{j,(t-1)}, &~\forall j\in\mathcal{J}, \,\forall t\in\mathcal{T}: t > 1  \label{phiB} \\
        &\phi_{j,t} \leq x_{j,t}, &~\forall j\in\mathcal{J}, \,\forall t\in\mathcal{T}  \label{phiC} \\
        &\phi_{j,t} \leq 2 - x_{j,t} - x_{j,(t-1)}, &~\forall j\in\mathcal{J}, \,\forall t\in\mathcal{T}: t > 1 \label{phiE} \\
        &\sum_{t=1}^{\textcolor{black}{w^{\min}_{j}}} x_{j,t} = 0,  & \forall j\in\mathcal{J} \, \label{window1} \\
        &\sum_{t=w^{\max}_{j}+1}^{T} x_{j,t} = 0, & \forall j\in\mathcal{J}  \label{window2}\\
        &\sum_{l=t}^{t+{t}^{\min}_{j}-1} x_{j,l} \geq {t}^{\min}_{j} \phi_{j,t},  &\forall t \in \{1,...,T-{t}^{\min}_{j} + 1\}, \forall j\in\mathcal{J} \label{c} \\
        &\sum_{l=t}^{t+{t}^{\max}_{j}} x_{j,l} \leq {t}^{\max}_{j},  &\forall t \in \{1,...,T-{t}^{\max}_{j}\}, \forall j\in\mathcal{J} \label{d} \\
        &\sum_{l=t}^{T} x_{j,l} \geq (T - t + 1) \phi_{j,t},  & \forall t \in \{T-{t}^{\min}_{j} + 2,...,T\}, \forall j\in\mathcal{J}\, \label{e} \\
        &\sum_{l=t}^{t+{p}^{\min}_{j}-1} \phi_{j,l} \leq 1,   & \forall t \in \{1,...,T-{p}^{\min}_{j}+1\}, \forall j\in\mathcal{J} \label{f} \\
        & \sum_{l=t}^{t+{p}^{\max}_{j}-1} \phi_{j,l} \geq 1,  & \forall t \in \{1,...,T-{p}^{\max}_{j}+1\},  \forall j\in\mathcal{J} \label{g} \\
        &\sum_{t=1}^{T} \phi_{j,t} \geq {y}^{\min}_{j}, &\forall j\in\mathcal{J}  \label{a} \\
        &\sum_{t=1}^{T} \phi_{j,t} \leq {y}^{\max}_{j}, &\forall j\in\mathcal{J}  \label{b} \\
        & \phi_{j,t}\in\{0,1\}, &~\forall j\in\mathcal{J}, t\in\mathcal{T}  \label{phi_bin} \\
        & x_{j,t}\in\{0,1\}, &~\forall j\in\mathcal{J}, t\in\mathcal{T}.  \label{x_bin}  
    \end{align}
\end{subequations}

Note that auxiliary binary variables \[
    \bm{\phi} = \left(\phi_{j,t}\right)_{\substack{j=1,\ldots,J \\ t = 1,\ldots,T}}
\] are used, which take a positive value if, and only if, job $j$ was not running at time $t-1$, but started running at time $t$.
Constraints \eqref{phiA} to \eqref{phiE} are used to ensure this behavior.
We also consider that jobs may run only during a time window defined by parameters $w^{\min}_{j}$ and $w^{\max}_{j}$, which is ensured by constraints \eqref{window1} and \eqref{window2}.
Such behavior is necessary to ensure that a payload, for instance, runs only when passing above a certain territory.

Jobs may also have limits on continuous execution.
If a job $j$ starts running at time $t$, then it must run for at least $t^{\min}_{j}$ time steps, and at most $t^{\max}_{j}$ time steps.
This is ensured by constraints \eqref{c} and \eqref{d}.
The formulation, through constraint \eqref{e}, also allows a job to start at the last time steps and keep running until the end, assuming it will keep running at the start of the following schedule.

A job may require to be executed periodically, at least every $p^{\min}_{j}$ time step, and at most every $p^{\max}_{j}$ time step.
This is ensured through constraints \eqref{f} and \eqref{g}, over the $\phi_{j,t}$ variables.
A job may also require multiple executions through the planning horizon, starting at least $y^{\min}_{j}$ times, and at most $y^{\max}_{j}$ times, which is ensured through constraints \eqref{a} and \eqref{b}.

Beyond job-specific restrictions in constraints \eqref{eq:qos-constraints}, the formulation also covers energy management,
\begin{subequations} \label{eq:energy-constraints}
\begin{align}
    &\sum_{j=1}^{J} q_{j} x_{j,t} \leq r_t + \gamma~V_{b}, & \forall t\in\mathcal{T} \label{EN_r} \\
    & b_{t} = r_{t} - \sum_{j \in \mathcal{J}} q_{j} x_{j,t}, &  \forall t \in \mathcal{T} \label{EN_b} \\
    & i_{t} = \frac{b_{t}}{V_{b}}, &  \forall t \in \mathcal{T} \label{EN_i}\\
    &\text{SoC}_{t+1} = \text{SoC}_{t} + \frac{i_{t}~e}{60~Q}, & \forall t \in \mathcal{T}  \label{EN_SOC1}\\
    &\text{SoC}_{t} \leq 1, & \forall t\in\mathcal{T} \label{EN_SOC2}   \\
    &\text{SoC}_{t} \geq \rho, & \forall t\in\mathcal{T} \label{EN_SOC3} 
.\end{align}
\end{subequations}
Parameter $r_t$ provides the power available from the solar panels at time $t$ and $q_j$ the power required for executing job $j$.
Thus, constraint \eqref{EN_r} limits the power consumption, with $\gamma \cdot V_b$ being the maximum power the battery can provide.
Constraints \eqref{EN_b} to \eqref{EN_SOC1} update the auxiliary variables $b_t$ and ${\rm SoC}_t$, which represent the exceeding power and State of Charge (SoC) at time $t$, based on the battery capacity $Q$ and the discharge efficiency $e$.
The SoC of the battery must stay within the limits given in constraints \eqref{EN_SOC2} and \eqref{EN_SOC3}, with $\rho$ being a lower limit usually greater than zero as a safety measure.

Thus, the MILP formulation is the maximization of \eqref{eq:qos} while subject to constraints \eqref{eq:qos-constraints} and \eqref{eq:energy-constraints}.
In other words, we can write any instance $I\in\mathcal{I}$, where $\mathcal{I}$ is the set of all possible instances of the ONTS problem, as
\begin{equation}\label{eq:formulation}
\begin{split}
    I : \max_{\bm{x},\bm{\phi},{\rm SoC}} ~& \underbrace{\sum_{j=1}^{J} \sum_{t=1}^{T} {u}_{j} x_{j,t} }_{ \text{QoS}}  \\
    \text{s.t.}  ~& \eqref{eq:qos-constraints},\eqref{eq:energy-constraints}  \\
    & \bm{x},\bm{\phi} \in \{0,1\}^{JT} , {\rm SoC}\in [0,1]^T
.\end{split}
\end{equation}
Note that the continuous variables ${\rm SoC}$ can be completely determined by the binary variables $\bm{x}$ and $\bm{\phi}$, so our problem can be reduced to finding an assignment $\bm{z}=(\bm{x},\bm{\phi}) \in \{0,1\}^{2JT}$.

Any instance $I\in\mathcal{I}$ is parameterized by the number of tasks $J$, the number of time units $T$ and the parameters $\bm{u}$, $\bm{q}$, $\bm{y}^{\min}$, $\bm{y}^{\max}$, $\bm{t}^{\min}$, $\bm{t}^{\max}$, $\bm{p}^{\min}$, $\bm{p}^{\max}$, $\bm{w}^{\min}$, $\bm{w}^{\max}$, $\bm{r}$, $\rho$, $e$, $Q$, $\gamma$, and $V_b$.
We will denote $\Pi_{J,T}$ the parameter space through which any instance $I\in\mathcal{I}$ can be uniquely determined by some parameter vector $\pi\in\Pi_{J,T}$ (given adequate $J$ and $T$).

\section{Graph Neural Networks (GNNs)}\label{sec:gnns}

Graph neural networks are generalizations of convolutional neural networks from grid-structured data to graphs.
The input to a GNN is a graph and a set of feature vectors associated with the graph nodes.
GNNs work by propagating feature information between neighboring nodes, which can be seen as message-passing.
More specifically, at each layer, the GNN first computes the messages sent by the nodes based on their feature vectors.
Then, it updates each node's feature vectors based on the messages sent by their neighbors.
We will refer to the feature update as a graph \emph{convolution}, due to its similarities to the operations in convolutional neural networks.

At each layer $l\in\{1,...,L\}$, the graph convolution operation has two major components: \textit{message functions} $M_l(\cdot)$, which compute the messages propagated by each node, and \textit{update functions} $U_l(\cdot)$, which update the node-associated features based on the messages.
Let $G = (V, E)$ be a graph and $\bm{h}^{(0)}_v$ be feature vectors for every node $v\in V$.
The messages sent $\bm{m}_{u}^{(l)}$ are computed based on the node features
\begin{equation}\label{eq:message-passing-messages}
    \bm{m}_{u}^{(l)} = M_l(\bm{h}_u^{(l-1)}), \forall u\in V
,\end{equation}
where $M_l(\cdot)$ is the message function of layer $l$.
After computing the messages, the features of node $v$ are updated through
\begin{equation}\label{eq:message-passing-updates}
    \bm{h}_v^{(l)} = U_l\left(\bm{h}_v^{(l-1)}, {\tt Aggregation}\left(\left\{\bm{m}^{(l)}_{u} : u\in \mathcal{N}(v) \right\}\right)\right)
,\end{equation}
where $\mathcal{N}(v)$ is the set of neighbors of $v$, $U_l(\cdot)$ is the update function of layer $l$, and ${\tt Aggregation}$ is a function that receives multiple message vectors and returns a single vector.

A common choice for graph convolutions is to use feedforward neural networks as message and update functions.
Given that the number of neighbors varies, we need an ${\tt Aggregation}$ function that always returns a fixed-length vector, invariant to input size, hence the naming.
A natural choice is to add up the messages, but many are the possibilities, such as
\begin{align}
{\tt Aggregation} = \begin{cases}
    \frac{1}{|\mathcal{N}(v)|} \sum\limits_{u \in \mathcal{N}(v)} \bm{m}_{u}^{(l)}, & \textrm{if mean} \\  
    \underset{u \in \mathcal{N}(v)}{\max}  \bm{m}_{u}^{(l)}, & \textrm{if max} \\   
    \sum\limits_{u \in \mathcal{N}(v)} \bm{m}_{u}^{(l)}, & \textrm{if sum} \\ 
    \sum\limits_{u \in \mathcal{N}(v)} \alpha_{u,v} \bm{m}_{u}^{(l)}, & \textrm{if attention} \\
\end{cases},
\end{align}
where $\alpha_{u,v}$ is an attention weight for node $u$ given node $v$, which can be learned along the model parameters.
Furthermore, the message function can easily be extended to consider the edge weight (or even edge features) along with the feature vectors of the neighbors.

A common approach is to define the message functions $M_l, l=1,\ldots,L$, as linear operators over the hidden features of the neighbors, aggregate these messages by summing, and use the ReLU (Rectifier Linear Unit) activation function with a bias as the update functions $U_l, l=1,\ldots,L$.
An example of such an approach is the model proposed by \citet{kipf_semi-supervised_2017}, which uses a linear combination of the features as the message function, aggregates messages by summation, and updates using a single-layer neural network with ReLU activation.
More precisely, we can describe their convolution operation as
\begin{equation}\label{eq:graph-conv}
\begin{aligned}
    \bm{m}_{u,v}^{(l)} &= \frac{1}{c_{vu}}W^{(l)}\bm{h}_u^{(l-1)},\, u\in \mathcal{N}(v) \\
    \bm{h}_{v}^{(l)} &= \text{ReLU}\left(\bm{b}^{(l)} + \sum_{u \in \mathcal{N}(v)} \bm{m}_{u,v}^{(l)}\right)
\end{aligned}
\end{equation}
where $c_{vu} = \sqrt{|\mathcal{N}(u)|}\sqrt{|\mathcal{N}(v)|}$ with $|\mathcal{N}(v)|$ denoting the number of neighbors, and $W^{(l)}\in \mathbb{R}^{d_l\times d_l},\bm{b}^{(l)}\in\mathbb{R}^{d_l}$ (layer $l$ has size $d_l$) are the weights and biases of the model, i.e., learnable parameters.

Another convolution operator was proposed by \citet{hamilton_inductive_2017} and named SAGE (SAmple and aGgrEgate).
The authors propose to directly aggregate the features of the neighbors, i.e., to use the identity as the message function and to use a single-layer network with ReLU activation as the update function.
Putting it into terms, the proposed graph convolution is
\begin{equation}\label{eq:sage-conv}
\begin{aligned}
    \bm{m}_{u,v}^{(l)} &= \bm{h}_u^{(l-1)},\, u\in \mathcal{N}(v) \\
    \bm{h}_{v}^{(l)} &= \text{ReLU}\left(\bm{b}^{(l)} + W^{(l)}_1 \bm{h}_v^{(l-1)} + W^{(l)}_2 {\tt Aggregation}\left (\bm{m}_{u,v}^{(l)},\, u\in \mathcal{N}(v)\right)\right)
\end{aligned}
\end{equation}
where $W^{(l)}_1,W^{(l)}_2\in \mathbb{R}^{d_l\times d_l},\bm{b}^{(l)}\in\mathbb{R}^{d_l}$ are the parameters.
The authors suggest using more complex aggregation operators, such as an LSTM (Long Short-Term Memory network) and a fully connected single-layer neural network, followed by a pooling operation (element-wise maximum).
Note that the SAGE convolution with summation as the aggregation function is equivalent to the one proposed by Kipf and Welling.

After recurrent graph convolutions through the $L$ layers of a GNN, we can use the resulting node features $H^{(L)} = \left[ \bm{h}_v^{(L)} \right]_{v\in V}$ straight-away, which is suitable for, e.g., node classification tasks, or we can aggregate them further into a single feature vector, suitable for, e.g., graph classification tasks.
The GNN can be trained end-to-end by minimizing a prediction loss based on its outputs, optimizing its parameters (e.g., $W^{(l)}$ and $\bm{b}^{(l)}$ of \eqref{eq:graph-conv}) in the same way as a traditional deep learning model.


\subsection{MILP Problems as Inputs for GNNs}\label{sec:instance-embedding}

We can feed MILP problem instances to GNNs by representing the relationship between problem variables and constraints through a bipartite graph~\citep{gasse_exact_2019,ding_accelerating_2020,khalil_mip-gnn_2022,nair_neural_2020,han_gnn-guided_2023}.
More precisely, given a problem of the form 
\begin{equation}\label{eq:opt}
\begin{aligned}
\max_{\bm{x}} & \quad \bm{c}^T \bm{x} \\
\text{s.t.:} & \quad A\bm{x} \le\bm{b} 
,\end{aligned}
\end{equation}
where $\bm{x}\in X \subseteq\mathbb{R}^n$, $A\in\mathbb{R}^{m \times n}$, $\bm{c}\in \mathbb{R}^n$, and $\bm{b}\in \mathbb{R}^m$, we can build a graph $G=(V_{\textrm{var}}\cup V_{\textrm{con}}, E)$, in which $|V_{\textrm{var}}| = n$, $|V_{\textrm{con}}|=m$, and $E=\{(v_{{\rm con},i},v_{{\rm var},j}) : A_{i,j} \neq 0\}$.
Intuitively, the graph represents the structure of the problem at hand, with edges capturing the relationship between variables and constraints.
Note that this approach yields a bipartite graph, that is a graph in which the nodes are separated into two disjoint sets, $V_{\textrm{var}}$ and $V_{\textrm{con}}$, with edges connecting only nodes from one set to the other.

To fully represent an MILP instance, however, the graph has to be enhanced with edge and node weights.
Let $w:V\cup E\mapsto \mathbb{R}$ be the weight function\footnote{Note that, in a slight abuse of notation, we use a single function to weigh both nodes and edges.}.
Then, for every node $v_i\in V_{\rm con}$, associated with the $i$-th constraint, the weight will be the constraint's bound, i.e., $i$-th element of $\bm{b}$, i.e., $w(v_i)=b_i$.
Similarly, for $v_j\in V_{\rm var}$, associated with the $j$-th variable, $w(v_j)=c_j$, where $\bm{c}=(c_j)_{j=1,\ldots,n}$.
Edge weights are given by the incidence matrix $A$, that is, given $e=(v_{con,i},v_{var,j}) \in E$, then $w(e)= A_{i,j}$.
This graph representation approach establishes a bijective relationship to the instance space, i.e., every MILP instance is uniquely determined by a weighted graph and vice-versa\footnote{For the weighted graph to identify an MILP instance uniquely, it would need to contain the information about the integrality of the variables. Restricting the bijection statement to MILP problems with the same number of integer variables would be more precise. Alternatively, to assume, without loss of generality, that the first $k\le n$ variables (i.e., $x_1$ to $x_k$) are integer variables and use $k$ as a weight for the whole graph.}.

In practice, the graph fed to a GNN is associated with feature vectors $\bm{h}_v^{(0)}, \forall v\in V$, of arbitrary size.
In other words, the information contained in the features provided to the network is a design choice.
It can contain the weights described above, but many other features might also help the model learn the graph-related task~\citep{gasse_exact_2019,nair_neural_2020}.

For illustration purposes, consider an optimization problem in the form of Eq. \eqref{eq:opt} with three variables $\bm{x} = [x_1, x_2, x_3]^T$, three constraints $C_1,C_2$ and $C_3$, and
\begin{equation}\label{eq:A-matrix-example}
    \bm{c} = \begin{bmatrix} 1 \\ 2 \\ 3 \end{bmatrix}; ~\\
    A = \begin{bmatrix} 1 & 2 & 0 \\ 0 & 1 & -1 \\ 5 & 0 & 1 \end{bmatrix}; ~\\
\bm{b} = \begin{bmatrix} 2 \\ 1 \\ 4 \end{bmatrix}
.\end{equation}
Then, a bipartite graph representation is illustrated in Figure \ref{fig:ilustra_b}.
For the features, one may define \[
\bm{h}_{x_j}^{(0)} = \begin{bmatrix}
    c_j \\
    \frac{1}{3}\sum_{i=1}^{3} A_{ij} \\
    \max_{i} A_{ij}
\end{bmatrix}\quad{\rm and}\quad\bm{h}_{C_i}^{(0)} = \begin{bmatrix}
    b_i \\
    \frac{1}{3}\sum_{j=1}^{3} A_{ij} \\
    \max_{j} A_{ij}
\end{bmatrix}
,\] such that \[
H^{(0)} = \begin{bmatrix}
    \bm{h}_{x_1}^{(0)T} \\
    \bm{h}_{x_2}^{(0)T} \\
    \bm{h}_{x_3}^{(0)T} \\
    \bm{h}_{C_1}^{(0)T} \\
    \bm{h}_{C_2}^{(0)T} \\
    \bm{h}_{C_3}^{(0)T}
\end{bmatrix} = \begin{bmatrix}
    1 & 2 & 5 \\
    2 & 1 & 2 \\
    3 & 0 & 1 \\
    2 & 1 & 2 \\
    1 & 0 & 1 \\
    4 & 2 & 5
\end{bmatrix}
.\]

\begin{figure}[!htb]
\centering
\begin{tikzpicture}
\node[circle,draw,fill =green!40] (x1) {$x_1$};
\node[ right = of x1,circle,draw,fill =green!40] (x2)  {$x_2$};
\node[ right = of x2,circle,draw,fill =green!40] (x3) {$x_3$};

\node[ below = of x1,circle,draw,fill =blue!40] (c1) {$C_1$};
\node[ below = of x2,circle,draw,fill =blue!40] (c2) {$C_2$};
\node[ below = of x3,circle,draw,fill =blue!40] (c3) {$C_3$};
\draw (x1) -- (c1);
\draw (x2) -- (c1);
\draw (x3) -- (c2);
\draw (x1) -- (c3);
\draw (x2) -- (c2);
\draw (x3) -- (c3);
\end{tikzpicture}
\caption{Bipartith graph representation of an MILP with three variables ($x_1$, $x_2$, and $x_3$), three constraints ($C_1$, $C_2$, and $C_3$), and $A$ as in Eq. \eqref{eq:A-matrix-example}.}\label{fig:ilustra_b}
\end{figure}

\section{Methodology}\label{sec:meth}

This section presents our methodological approach to tackle the two research questions proposed in the introduction.
As both questions concern learning tasks with GNNs on to the ONTS problem, we describe our approach to generating problem data used to train the deep learning models, and our proposed GNN architecture (SatGNN). 
Furthermore, and specifically for the experiments that tackle the second research question, we show our approach to embed trained deep learning models in heuristic solutions for the ONTS problem.

\subsection{Data}\label{sec:data}

High-quality data is necessary to learn the tasks of interest and perform reliable evaluations.
To achieve a significant quantity of data, we generate new instances of the ONTS problem with variations of all task parameters and energy availability.
However, randomly sampling parameter values suitable for the problem formulation presented in Sec. \ref{sec:problem} would not reflect the distribution of instances we expect to see in practice.
Therefore, we use a methodological approach to sample the parameter values, following \citet{rigo_instance_2023}.
The procedure for generating pseudo-random realistic parameters is detailed in \ref{appx:random-instance}.

The algorithm for constructing our dataset is formalized in Algorithm \ref{alg:dataset-generation}.
We optimize every new instance $I\in\mathcal{I}$ using a commercial solver with limited time, during which we collect the best candidate solutions found (represented by $Z^\star$).
Note that we collect only the values for the binary variables $\bm{z}$, as the binary ones can completely determine the continuous variables (see Sec. \ref{sec:problem}).
The instance is rejected if no feasible solution is found during the time budget (or if the solver proves infeasibility).

\begin{algorithm}[h]
    \SetAlgoLined
    \KwData{Time horizon $T$, number of jobs $J$, number of instances (final dataset size) $n$.}
    \KwResult{Dataset $\mathcal{D} = \{(I,Z^\star): Z^\star\subset Z\,{\rm given}\,I\}$.}
    
    \While{$|\mathcal{D}| < n$}{
        $\pi \gets {\tt RandomInstance}(T,J)$ \\
        $I \gets {\tt BuildInstance}(\pi)$ \\
        $Z^\star \gets {\tt Solver}(I)$
        
        \If{$|Z^\star| > 0$}{%
            $\mathcal{D}$.add$(I, Z^\star)$
        }
    }
    \caption{Dataset generation. $\pi$ is the parameter vector described in Sec. \ref{sec:problem}, $Z$ is the set of all feasible solutions, and $Z^\star \subset Z$ is the set of candidate solutions the solver finds.
    For a description of ${\tt RandomInstance}$ see \ref{appx:random-instance}.}\label{alg:dataset-generation}
\end{algorithm}

\subsection{SatGNN}\label{sec:sat-gnn}

\begin{figure}[h]
    \centering
    \includegraphics[width=0.99\textwidth]{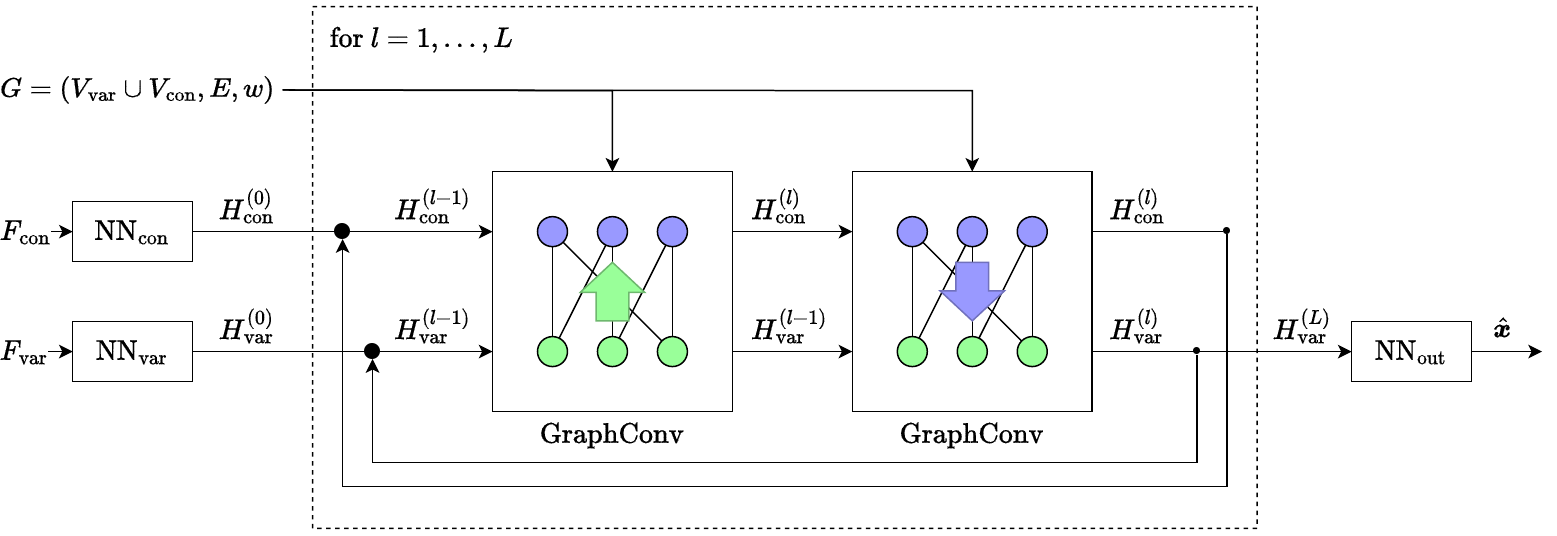}
    \caption{\emph{SatGNN} architecture overview, where $G$ is the bipartite graph representation of an MILP problem instance, $H$ variables represent the sets of hidden features of the nodes, and $\rm{NN}_\cdot$ are neural networks. The connection of $G$ and the $GraphConv$ operators represents both the weights of the edges and the neighborhood information.}
    \label{fig:sat-gnn-overview}
\end{figure}

We name \emph{SatGNN} the base model proposed for all experiments.
An overview of the components of our model can be seen in the diagram of Figure \ref{fig:sat-gnn-overview}, \new{and a summary of the symbols used is presented in \ref{appx:gnn-symbols}}.

\new{%
SatGNN is designed to take as input instances of the ONTS problem.
Each instance is a mathematical program with a varying number of variables and constraints (i.e., varying number of jobs).
Thus we expect our model to handle inputs of different sizes and to be invariant to the symmetries that mathematical programs exhibit (e.g., permutations of variables, constraints). 
For that reason, we follow the approach described in Section~\ref{sec:gnns} of embedding each instance as a bipartite graph, and design SatGNN around a GNN (hence its name).
}


\new{To fully represent each instance, the input graph is augmented with feature vectors for every node.}
Let $G = (V, E, w)$ be the graph representation of an instance $I \in \mathcal{I}$, as described in Sec.~\ref{sec:instance-embedding}, with $V = V_{\textrm{var}} \cup V_{\textrm{con}}$ being the set of nodes and $w: E \longrightarrow \mathbb{R}$ the edge-weight function based on the variable coefficients in the constraints.
Let $\bm{f}_v$ denote the feature vector associated with node $v$.
We add four features to constraint nodes and six to variable nodes, as described in Table~\ref{tab:feature-desc}.
\new{We highlight that the features overdetermine the instance; that is, more features than necessary are provided to the model. This is common in deep learning practice and has been observed to yield positive results in combinatorial optimization applications~\citep{nair_neural_2020,khalil_mip-gnn_2022}.}

\begin{table}[h]
    \centering
        \caption{Description of input features for SatGNN.}
    \label{tab:feature-desc}
    \begin{tabular}{p{6.5cm}|p{6.5cm}}
    \toprule
        Features of constraint nodes ($\bm{f}_{v_{\rm con}}$) & Features of variable nodes ($\bm{f}_{v_{\rm var}}$) \\
    \midrule
         Constraint's upper bound ($\bm{b}$)                     &  Variable's coefficient in the objective ($\bm{c}$)\\
         Constraint's average coefficient (mean of $A_{i*}$)     &  Variable's average coefficient in the constraints (mean of $A_{*j}$) \\
         Number of neighbors/non-null coefficients ($|\mathcal{N}(v_{\rm con})|$)    &  Number of neighbors/non-null coefficients ($|\mathcal{N}(v_{\rm var})|$) \\
         Whether it is an equality or an inequality constraint &  Largest coefficient in the constraints ($\max(A_{*j})$) \\
                                                                    &  Smallest coefficient in the constraints ($\min(A_{*j})$) \\
                                                                    &  Whether it is a continuous or binary variable \\
    \bottomrule
    \end{tabular}
\end{table}

The feature vectors are encoded into the first layer of hidden features by fully-connected neural networks.
More precisely, the hidden feature vectors $H^{(0)}\in \mathbb{R}^{(n+m)\times d}$ (where $n$ is the number of variables, $m$ is the number of constraints, and $d$ is the dimension of the hidden feature vectors) are computed by single-layer neural networks with ReLU activation
\begin{align*}
    {\rm NN}_{\rm var}:\mathbb{R}^6& \longrightarrow\mathbb{R}^{d}_+ \\
    \bm{f}_{v_{\rm var}} &\longmapsto \bm{h}^{(0)}_{v_{\rm var}} = {\rm NN}_{\rm var}(\bm{f}_{v_{\rm var}})
\end{align*}
and
\begin{align*}
    {\rm NN}_{\rm con}:\mathbb{R}^4& \longrightarrow\mathbb{R}^{d}_+ \\
    \bm{f}_{v_{\rm con}} &\longmapsto \bm{h}^{(0)}_{v_{\rm con}} = {\rm NN}_{\rm con}(\bm{f}_{v_{\rm con}})
,\end{align*}
such that the initial hidden features associated to the variable nodes $H^{(0)}_{\rm var} \in \mathbb{R}^{n\times d}$ are computed through ${\rm NN}_{\rm var}$ and the initial hidden features associated to the constraint nodes $H^{(0)}_{\rm con} \in \mathbb{R}^{m\times d}$ are computed through ${\rm NN}_{\rm con}$, in which $H^{(0)} = [H^{(0)}_{\rm var},H^{(0)}_{\rm con}]$.

At each layer of SatGNN, the graph convolutions are performed in two steps, as in \citet{gasse_exact_2019}.
First, a graph convolution is applied to update the hidden features of the constraint nodes.
Then, another graph convolution is performed, this time considering the updated features of the constraint nodes to update the hidden features of the variable nodes.
These two operations are illustrated in Figure \ref{fig:sat-gnn-overview} through the $GraphConv$ blocks.

Finally, after $L$ iterations, the last layer's hidden features associated to the variable nodes are fed to a neural network
\begin{align*}
    {\rm NN}_{\rm out}:\mathbb{R}^d_+& \longrightarrow\mathbb{R} \\
    \bm{h}^{(L)}_{v_{\rm var}} &\longmapsto \hat{z}_{v_{\rm var}} = {\rm NN}_{\rm out}(\bm{h}^{(L)}_{v_{\rm var}})
.\end{align*}
This network combines the features extracted by the graph convolution operations into the desired shape of the output.
The network contains two layers of $d$ nodes with ReLU activation.
For node classification tasks (Sec. \ref{sec:meth-sol-pred}), the output of the ${\rm NN}_{\rm out}$ contains a sigmoid activation function, resulting in a probability prediction for each variable node.
For graph classification tasks (Sec. \ref{sec:meth-feas-classification}), in which the output is a single probability value for the entire graph, the output of ${\rm NN}_{\rm out}$ for every variable node is summed into a single value before the sigmoid.

\subsection{Training}\label{sec:training}

Two learning tasks are proposed to investigate the research questions that are the focus of this paper. To investigate whether a GNN can learn the structure of the ONTS problem (\textit{first research question}), we train the SatGNN model on the feasibility classification of candidate solutions. To evaluate the effectiveness of a GNN-based heuristic (\textit{second research question}), we train SatGNN to generate candidate solutions given instances of the problem.

\subsubsection{Feasibility Classification}\label{sec:meth-feas-classification}

\new{%
Given an instance $I\in\mathcal{I}$ of the ONTS problem and a candidate solution $\bm{\hat{z}}$, we aim for the model to classify whether  $\hat{\bm{z}}$ is a feasible solution. 
We address this task using SatGNN by embedding the candidate solution as an additional feature in each variable node, thereby adding one extra dimension to every $\bm{f}_{v_\text{var}}$. 
Also, the model’s output is a binary value (0 or 1), indicating whether the assignment is infeasible or feasible, respectively.
}

As briefly described in Section \ref{sec:sat-gnn}, the feasibility classification is a graph classification task.
\new{%
It is therefore natural to take advantage of maximum likelihood training.
To this end, we define $y=p(\hat{\bm{z}}\in Z | I) \in \{0,1\}$ as the target for each assignment and instance.
Note that $p(\hat{\bm{z}}\in Z | I)$, which represents the probability that assignment $\hat{\bm{z}}$ is feasible, matches the label, once there is no stochastic component in the feasibility.
We then use SatGNN to obtain an estimate $\hat{p}(\hat{\bm{z}} \in Z \mid I) \in [0,1]$ of the true probability.
In particular, we do so by aggregating the output of the last fully-connected module ${\rm NN}_{\rm out}$, and using the sigmoid function.
}


\new{%
The model is trained to minimize the binary cross-entropy (BCE) between the predicted probability $\hat{p}(\hat{\bm{z}}\in Z | I)\in[0,1]$ and the true probability $p(\hat{\bm{z}}\in Z | I)\in[0,1]$.
We note that the minimization is performed over the parameters of the graph convolutions, as well as the parameters of ${\rm NN}_{\rm con}$, ${\rm NN}_{\rm var}$, and ${\rm NN}_{\rm out}$.
}

The training data is built with instances from $\mathcal{D}$ (see Sec. \ref{sec:data}).
\new{%
However, in each  $(I,Z^\star)\in\mathcal{D}$, we only have high-quality feasible solutions in $Z^\star$, which is clearly not representative of the input space for the feasibility classification task. 
In other words, we need both feasible and infeasible candidate solutions for the model to learn the classification.
}

\new{%
For that end, we augment  $Z^\star$ with two other sets of candidate solutions.  
The first, which we call $Z_R$, is constructed from candidate solutions uniformly sampled from $\{0,1\}^{2JT}$, forming a representative sample of the entire domain.  
The second set, $Z_N$, is constructed from candidate solutions in the neighborhood of $Z^\star$.  
Our intuition behind $Z_N$ is to increase the information density of our sample near the borders of the feasible region, which is where our target function is highly nonlinear.  
To achieve this, we build $Z_N$ by randomly flipping a limited number of assignments in randomly selected candidate solutions from $Z^\star$.
}

\new{%
Finally, our dataset for feasibility classification can be described as \[
    \mathcal{D}_{\rm feas} = \left\{ (I,\hat{\bm{z}},y)\in \mathcal{I} \times Z\times \{0,1\} : (I,Z^\star) \in \mathcal{D}, \hat{\bm{z}}\in Z^\star\cup Z_R\cup Z_N , y = p(\hat{\bm{z}}\in Z | I) \right\}
,\]
\new{where $p(\hat{\bm{z}}\in Z | I)$ is computed for each candidate solution\footnote{Except for those in $Z^\star$, which are always feasible.} by evaluating the constraints of $I$.}
}



\subsubsection{Optimal Solution Prediction}\label{sec:meth-sol-pred}

\new{%
We want SatGNN to predict an optimal solution given an instance of the ONTS problem.
We expect this task to be more challenging than the previous one, as finding an optimal solution is significantly harder than checking feasibility.
Furthermore, an instance may have multiple optimal solutions, so the target for the model is not trivially well-defined.
}

\new{%
We formulate this task through the variable bias at optimality, that is, the probability that each binary variable takes a positive value in an optimal solution chosen at random under a uniform distribution.
We denote this probability vector as \[
\bm{p}(\bm{z}^\star = 1 | I) = \begin{bmatrix}
     p(z^\star_{1,1} = 1 | I) \\ \vdots \\ p(z^\star_{J,T} = 1 | I)]
\end{bmatrix}
,\]
where $p(z^\star_{j,t} = 1 | I)$ is the probability that the variable $z^\star_{j,t}$ takes a positive value in an optimal solution to instance $I$.
We use SatGNN to obtain an estimate $\hat{\bm{p}}(\bm{z}^\star = 1 | I)$, which we can sample to generate candidate solutions for $I$.
In particular, we use thresholding to obtain a candidate solution.
}


\new{%
The standard classification approach for this task is to train the model to minimize the BCE with respect to the best-known solution, which serves as the target.
   Note that the best-known solution is indeed the target when the model has a single optimal solution and the instance has been solved to optimality.
For training with the \emph{best solution}, we build the dataset by selecting the best $\bm{z}^\star\in Z^\star$ for each $(I,\bm{z}^\star) \in\mathcal{D}$, i.e., \[
    \mathcal{D}_\text{opt-b} = \left\{ (I,\bm{z}^\star) \in \mathcal{I}\times Z : (I,Z^\star)\in \mathcal{D}, \bm{z}^\star = \arg\max_{\hat{\bm{z}}\in Z^\star} {\rm QoS}(\hat{\bm{z}}; \bm{u}_I) \right\}
,\] in which $\bm{u}_I$ is the vector of priorities from instance $I$ (see Sec. \ref{sec:problem}).
}

Another training approach is the one proposed by \citet{nair_neural_2020}, in which multiple known solutions are used as targets, being weighted by their objective value.
Note that the best-known solution is also used in this approach, alongside sub-optimal solutions.
\new{Intuitively, this \emph{multi-target} approach guides the model toward the region of high-quality solutions, rather than solely toward the best-known solution, which may result in more feasible predictions.}

\new{%
The dataset for the multi-target approach simply requires computing the weights for each $\hat{\bm{z}}\in Z^\star$ of each instance in $\mathcal{D}$.
More specifically, we build
\[
    \mathcal{D}_\text{opt-m} = \left\{ (I,\hat{\bm{z}},w_{\hat{\bm{z}}}) \in \mathcal{I}\times Z : (I,Z^\star)\in \mathcal{D}, \hat{\bm{z}}\in Z^\star, w_{\hat{\bm{z}}} = \frac{\exp({\rm QoS}(\hat{\bm{z}}; \bm{u}_I))}{\sum_{\bm{z} \in Z^\star}\exp({\rm QoS}(\bm{z}; \bm{u}_I))}  \right\}
.\]
}

\subsection{SatGNN-based Heuristic}\label{sec:meth-heuristics}

\new{%
Although we could use the SatGNN by itself as a heuristic for the ONTS problem by sampling its output to generate candidate solutions, it is unlikely that these solutions will be useful.  
Because our problem has constraints, an error in a single variable may result in an infeasible candidate. 
}
Given that ONTS problems of moderate size already have thousands of variables\footnote{The smallest instance evaluated by \citet{rigo_branch-and-price_2022} has 1746 binary variables, considering both $x$ and $\phi$ variables.}, infeasibility is likely to occur.
\new{%
To circumvent this limitation, we use the deep learning model in conjunction with a mathematical programming solver as a primal heuristic.
In particular, we propose three approaches for doing so, which instantiate recent GNN-guided baselines~\citep{nair_neural_2020,han_gnn-guided_2023}.}

\subsubsection{Warm-starting}\label{sec:meth-heuristics-ws}

\new{%
A straightforward way to aid most commercial-grade solvers is to warm-start them with good candidate solutions.
}
For example, the SCIP solver~\citep{bestuzheva_scip_2021} accepts complete and partial solutions to the problem, which are used to guide the inner heuristics during the optimization process.
We use the output of the model to determine which variables will compose the partial solution provided to the solver based on the \emph{confidence} of the model's prediction.
\new{In other words, we warm-start the solver with partial solutions built using only the assignments for variables for which the model predicts values closest to 0 or 1.}

Formally, let $\hat{\bm{p}}(\bm{z}^\star = 1|I)$ be the output of SatGNN given an instance $I\in\mathcal{I}$ as defined in Sec. \ref{sec:meth-sol-pred}.
\new{%
Our definition of confidence for a given variable comes from the maximum predicted marginal probability.
For instance, the confidence with respect to variable $x_{j,t}$ of an instance $I$ is
}
\[
\kappa(x_{j,t}|I) := \begin{cases}
    \hat{p}(x_{j,t}^\star = 1|I), &\text{if}\quad\hat{p}(x_{j,t}^\star = 1|I) \ge 0.5 \\
    \hat{p}(x_{j,t}^\star = 0|I) = 1 - \hat{p}(x_{j,t}^\star = 1|I), &\text{otherwise}
.\end{cases}
\] The confidence associated with the $\phi_{j,t}$ variables can be described similarly.
Note that the confidence is always between 0.5 and 1.
\new{Furthermore, we write $\bm{\kappa}(\bm{z}^\star|I)$ to denote the confidence vector over all variables.}

Let us write $\bm{z}=(\bm{x},\bm{\phi}) = (z_1, \ldots, z_k,\ldots, z_{2JT})$.
We define $\mathcal{C}^{N}_{I}\subseteq \{1,\ldots,2JT\}$ as the set of indices of the $N$ variables that the model is most confident of, that is, $|\mathcal{C}^{N}_I| = N$ and \[
    \mathcal{C}^{N}_I = \left\{ k \in \{1,\ldots,2JT\} : \kappa(z^\star_k|I) \ge \kappa(z^\star_{k'}|I), \forall k' \not\in \mathcal{C}^{N}_I \right\}
.\] Then, a partial solution $\bar{\bm{z}}^{(N)}$ of size $N$ can be written \[
    \bar{\bm{z}}^{(N)} = \left[\hat{z}_{k}\right]_{k\in \mathcal{C}^{N}_{I}}
,\] which contains the predicted assignments of the highest confidence and can be provided to the solver.

Note that warm-starting does \emph{not} configure a heuristic solution to the problem, as it just modifies the behavior of the heuristics already present in the algorithmic solution.
Therefore, we maintain optimality and feasibility guarantees, even in the worst-case scenario when the model provides the solver with bad, infeasible solutions.

\subsubsection{Early-fixing}\label{sec:meth-heuristics-ef}

\new{%
Our second matheuristic is similar to the Neural Diving approach proposed by \citet{nair_neural_2020}.
Instead of merely informing the solver of a partial candidate solution via warm-starting, we modify the instance of the problem and fix the values of the variables that the model is most confident of.
}
More precisely, early-fixing is equivalent to adding constraints of the form \[
    z_{k} = \hat{z}_{k}, \forall k\in \mathcal{C}^{N}_{I}
\] to the mathematical program of instance $I$, limiting $N$ variables to the assignments predicted by the model.
In practice, however, no constraints are added, \new{but the variables are replaced by the assignment in the model formulation.}
\new{This is equivalent to a diving heuristic with respect to the branch-and-bound tree of instance $I$.}
The resulting sub-problem will contain fewer binary variables and, thus, is expected to be faster to solve.

\new{%
We highlight, however, that while \citet{nair_neural_2020} sample multiple partial solutions from the model output and solve all of them, we sample only one, and with a fixed size.
This design choice is based on the fixed time-budget available for the ONTS problem at hand (see Section~\ref{sec:onts}).
With a single partial solution of fixed size, we have more control over the amount of binary variables in the program fed to the solver, which correlates to the solving time.
In other words, we trade solution quality, which would be expected to be higher if more partial solutions were evaluated, for lower computational cost, since a single problem is solved.
}

Note that every feasible solution to the early-fixed problem and the fixed variables is also feasible in the original instance.
However, an optimal solution to the sub-problem might not be optimal for the original instance.
Even worse, early-fixing the problem might render it infeasible if the \new{model's output result in an infeasible partial assignment.}
Therefore, this approach of early-fixing is a matheuristic, as it uses mathematical optimization to find heuristic solutions.

\subsubsection{Trust Region}\label{sec:meth-heuristics-tr}

\new{%
The third approach follows the learning-based heuristic proposed by \citet{han_gnn-guided_2023}.
}
Instead of fixing the variables that the model is most confident of, \new{we allow a limited deviation from the sampled partial candidate solution.}
In other words, we define a \emph{trust region} around the candidate solution.

\new{In practice, a new constraint is added to the mathematical programming formulation of each instance $I$ that limits feasible solutions to a neighborhood around the partial solution predicted by SatGNN.}
The constraint has the form \[
    \sum_{k\in \mathcal{C}^N_I} \left|z_{k} - \hat{z}_{k} \right| \le \Delta
,\] where $\Delta \in \mathbb{N}$ is the size of the trust region.

\new{%
Note that we do not make any distinctions between 1 and 0 assignments, as done in \citet{han_gnn-guided_2023}.
We follow the same confidence-based approach as in the previously proposed methods to select the variables.
We do so because there is no \emph{a priori} bias in optimization problems toward 1 or 0 values.  
Furthermore, even if the instances in $\mathcal{I}$ exhibit such a bias---as is indeed the case for the ONTS problem---we expect the model to learn it during training, so that it is reflected in the predicted marginal probabilities (and, consequently, in the confidence score).
}

In the above definition, it is easy to see that the early-fix is a particular case of the trust region method with $\Delta=0$.
In fact, just like the early-fixing approach, allowing a trust region around the (partial) candidate solution also configures a matheuristic, as neither optimality nor feasibility are guaranteed.

\section{Experiments and Results}\label{sec:exps}

The experiments for this paper were conducted in Python, using PyTorch, the DGL libraries, and the SCIP solver, on a server with an Intel i7-12700  16-Core (12 cores, 20 threads), an NVIDIA RTX A4000, 16 GB of memory, running Ubuntu 22.04.1 LTS 64 bits.
All the code written for our experiments, as well as detailed instructions on how to reproduce our results, are available in the paper's accompanying repository\footnote{\url{https://github.com/brunompacheco/sat-gnn}}.

In the following sections, we present three experiments addressing the two research questions raised in the introduction (Sec. \ref{sec:intro}).
In the first experiment, we train the proposed SatGNN model to classify the feasibility of candidate solutions for problem instances.
We consider three scenarios of increasing difficulty with respect to the generalization performance.
In the second experiment we train SatGNN models to predict the bias for the binary variables of problem instances, effectively learning the probability that each binary variable has to assume a positive value in an optimal solution.
We compare two training approaches, considering either the optimal solution as the target or multiple feasible solutions as targets for each instance.
Finally, we use the models trained for optimality prediction to generate candidate solutions, which are used to construct heuristics to the ONTS problem.
More specifically, we use the trained models to construct three approaches (as presented in Sec. \ref{sec:meth-heuristics}) to improve the SCIP solver in its off-the-shelf setting: warm-starting, early-fixing, and trust region.

\subsection{Dataset}\label{sec:exp-datasets}

The data for the following experiments is generated according to the procedure described in Sec. \ref{sec:data}.
The dataset construction requires verifying feasibility solving to (quasi-)optimality instances of the ONTS problem.
In other words, the dataset construction imposes a high computational cost, which scales with the size of the problem.

To circumvent the cost of dataset generation, we build a training set with many smaller instances and validation and test sets with fewer but larger instances.
This way, we can afford a sufficient volume of data for training while evaluating at the most relevant instances, which have a higher solving cost.
At the same time, we test the model's generalization capability by evaluating its performance on instances harder than those seen during training.

We base our data generation in the instances proposed by \citet{rigo_instance_2023}, which take the FloripaSat-I mission as a reference~\citep{marcelino_critical_2020} (see \ref{appx:random-instance}).
More precisely, we focus on instances with scheduling horizon $T=125$.

The dataset is generated through Algorithm \ref{alg:dataset-generation}.
We solve each instance and gather 500 of the best solutions possible, limited to a time budget of 5 minutes.
More precisely, every instance $(I, Z^\star)\in \mathcal{D}$ is such that $|Z^\star| = 500$.
An example of the optimal schedule for one of the generated instances with $J=9$ can be seen in Figure \ref{fig:example-scheduling}.
All instances and solutions are made publicly available~\citep{pacheco_bruno_m_2023_8356798}.

We generate 200 instances for each $J\in\{9,13,18\}$ jobs, and 40 instances for each $J\in\{20,22,24\}$ jobs, in a total of 720 instances.
A summary of the size (in terms of variables and constraints) of each instance in our dataset is seen in Table~\ref{tab:dataset-description}.
We underscore the difference in difficulty between the instances in the training set and the ones in the validation and test sets, as evidenced by the significant increase in the average primal-dual gap and decrease in the lower bound as measured by the relative objective value, when solving those instances using our baseline method (SCIP) within the time limit.

\begin{table}[h]
    \centering
    \caption{%
    Summary of instances in our dataset.
    Columns $m$ and $n$ indicate the number of constraints and variables, resp., in each instance.
    Under \emph{Dataset}, we present the number of instances in each part of our dataset.
    Columns in \emph{Baseline optimization results} have summary statistics after using our baseline optimization method (SCIP) with the time budget ($T$).
    More specifically, \emph{Feasible} and \emph{Optimal} indicate, respectively, the fraction of instances that were solved to feasibility and optimality under the time budget, \emph{Gap} is the primal-dual gap, and \emph{LB} is the average relative objective of the candidate solutions.
    }
    \label{tab:dataset-description}
    \begin{tabular}{ccc|ccc|cccc}
    \toprule
               &       &      & \multicolumn{3}{c|}{Dataset} & \multicolumn{4}{c}{Baseline optimization results} \\
    \# of jobs & $m$   & $n$  & Train & Val. & Test          & Feasible & Optimal & Gap  & LB \\
    \midrule
    9          & 7518  & 2250 & 200   &      &               & 98\%     & 20\%    & 0.03  & 98\%      \\
    13         & 10618 & 3250 & 200   &      &               & 96\%     & 3\%     & 0.05  & 95\%      \\
    18         & 13434 & 4500 & 200   &      &               & 85\%     & 0\%     & 0.09  & 83\%      \\
    20         & 14633 & 5000 &       & 20   & 20            & 78\%     & 0\%     & 0.13  & 75\%      \\
    22         & 16430 & 5500 &       & 20   & 20            & 60\%     & 0\%     & 0.17  & 56\%      \\
    24         & 18664 & 6000 &       & 20   & 20            & 48\%     & 0\%     & 0.31  & 42\%      \\
    \bottomrule
    \end{tabular}
\end{table}

\subsection{Feasibility Classification}\label{sec:exp-feas}

As described in Sec. \ref{sec:meth-feas-classification}, we build the dataset for feasibility classification $\mathcal{D}_{\rm feas}$ by adding random candidate solutions to $Z^\star$ for every $(I,Z^\star) \in \mathcal{D}$.
More precisely, for every instance of the ONTS problem, we generate $Z_R$ and $Z_N$ such that $|Z_R|=250$ and $|Z_N| = 250$, in a total of 1000 solutions for every instance.

SatGNN is modified for graph classification as described in Sec. \ref{sec:meth-feas-classification}.
We use the graph convolution proposed by \citet{kipf_semi-supervised_2017}, as presented in equation \eqref{eq:graph-conv}, and a single layer ($L=1$).
The hidden features all have a fixed size $d=8$.
Both ${\rm NN}_{\rm con}$ and ${\rm NN}_{\rm var}$ are neural networks with a single layer and ReLU activation, while ${\rm NN}_{\rm out}$ is an artificial neural network with 3 layers and ReLU activation in the hidden layers and sigmoid activation at the last layer.
For all feasibility classification experiments, SatGNN is trained using Adam~\citep{kingma_adam_2015} with a learning rate of $10^{-3}$ to minimize the binary cross-entropy between the prediction and the feasibility of the candidate solution.
The model is trained until the training loss becomes smaller than $10^{-2}$, limited to a maximum of 200 epochs.

First, SatGNN is trained for one particular instance.
Given an instance $I$, we build $\mathcal{D}_{\rm feas}^{I} = \{ (\bm{\hat{z}},y)\in Z\times \{0,1\} : (I,\bm{\hat{z}},y)\in\mathcal{D}_{\rm feas}\}$.
This dataset is randomly divided into a training and a test set in an 80-20 split, so 800 instances are used to train the model.
For each number of jobs $J\in\{9,13,18,20,22,24\}$, we select 5 different instances randomly.
In all experiments (all instances of all sizes), \emph{SatGNN achieves 100\%  accuracy}, that is, the model is able to perfectly distinguish between feasible and infeasible candidate solutions in the test datasets.

\begin{figure}
    \centering
    \includegraphics{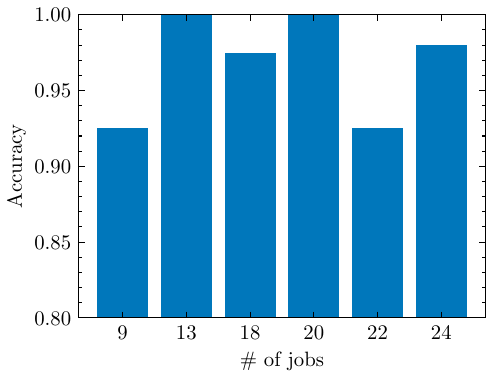}
    \caption{Test set performance of SatGNN trained for feasibility classification of candidate solutions given a fixed number of jobs. All instances used for testing were not seen by the model beforehand.}
    \label{fig:across-instances-accuracy}
\end{figure}

Then, SatGNN is trained to generalize across instances with the same number of jobs.
For each number of jobs $J$, we build the dataset by selecting only instances (and the respective candidate solutions) with the same number of jobs, i.e.,
\begin{equation}\label{eq:D-feas-jobs}
    \mathcal{D}_{\rm feas}^{J} = \{(I,\hat{\bm{z}},y)\in \mathcal{D}_{\rm feas} : J_I = J \}
,\end{equation}
where $J_I$ represents the number of jobs of instance $I$.
20 of the generated instances are randomly selected for training, in a total of 20,000 training samples, while 10 different instances are randomly selected for testing.
SatGNN achieves a very high performance, with over 90\% accuracy on all settings.
The complete test set performance for the different number of jobs can be seen in Figure \ref{fig:across-instances-accuracy}.

Finally, SatGNN is trained solely on candidate solutions from small instances (i.e., those with $J\in\{9,13,18\}$) and tested on candidate solutions from large instances (i.e., those with $J\in\{20,22,24\}$).
Following the notation of equation \eqref{eq:D-feas-jobs}, we use $\mathcal{D}_{\rm feas}^9\cup \mathcal{D}_{\rm feas}^{13}\cup\mathcal{D}_{\rm feas}^{18}$ for training and $\mathcal{D}_{\rm feas}^{20}\cup \mathcal{D}_{\rm feas}^{22}\cup\mathcal{D}_{\rm feas}^{24}$ for testing.
In total, 60,000 samples are used for training and 30,000 samples are used for testing.
SatGNN achieves 94.15\% accuracy in the test set.
The performance across the different sizes of instances and the groups of candidate solutions is presented in Table \ref{tab:across-sizes-accuracy}.
We see that the model has more difficulty in the candidate solutions from the $Z_N$ sets, the candidate solutions in the edge of the feasible region.
It is also true that the performance decreases with problem size (number of jobs), indicating an increasing difficulty, as expected.

\begin{table}[h]
    \centering
    \caption{Feasibility classification performance of SatGNN on instances larger than those used for training. The test set is discriminated into the three sets of candidate solutions that compose it (see Sec. \ref{sec:meth-feas-classification}). Each row indicates the set of 10 instances with the same size (number of jobs).}
    \label{tab:across-sizes-accuracy}
    \begin{tabular}{c | cccc}
    \toprule
               & \multicolumn{4}{c}{Accuracy}          \\
    \# of jobs & $Z^\star$ & $Z_N$ & $Z_R$   & Total   \\
    \midrule
    20         & 100\%     & 90\%  & 100\%   & 97.5\%  \\
    22         & 100\%     & 80\%  & 100\%   & 95\%    \\
    24         & 90\%      & 90\%  & 89.84\% & 89.96\% \\
    \bottomrule
    \end{tabular}
\end{table}

\begin{figure}
    \centering
    \includegraphics{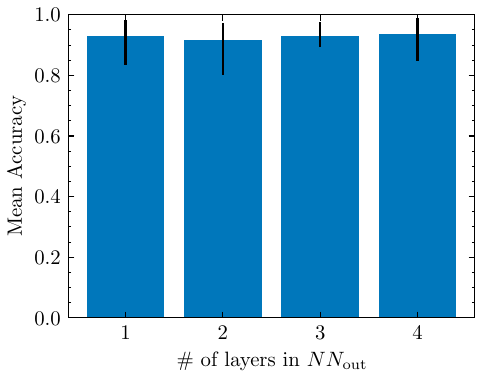}
    \caption{%
    Test set performance of SatGNN with a varying number of layers in $\rm{NN}_{\text{out}}$. Accuracy is measured as an average over 5 random repetitions of the experiment. Vertical bars indicate the minimum and maximum values observed.
    }
    \label{fig:nnout-layers-ablation-accuracy}
\end{figure}

To further assess the architecture of the SatGNN model, we perform an ablation study on the number of layers in the ${\rm NN}_{\text{out}}$ component.
We repeat the training and evaluation of the SatGNN model in the feasibility classification task across instance sizes for different numbers of layers in the last MLP of our model.
More specifically, we experiment with up to 4 layers in ${\rm NN}_{\text{out}}$, and for each number of layers, we repeat the experiment 5 times with different random initializations.
The average performance of those models appears in Figure~\ref{fig:nnout-layers-ablation-accuracy}.
No significant difference can be observed between the experiments, indicating that the learning primarily occurs before ${\rm NN}_\text{out}$, in the graph convolutional layers.
In fact, applying the Wilcoxon signed-rank test~\citep{wilcoxon_1945} shows no statistically significant ($p$-value$>0.05$) difference between the samples; however, this result should be taken with caution in the face of the size of the samples.

\subsection{Optimal Solution Prediction}\label{sec:exp-opt-pred}

Following the methodology presented in Sec. \ref{sec:meth-sol-pred}, we first generate the datasets $\mathcal{D}_{\rm opt\text{-}b}$ and $\mathcal{D}_{\rm opt\text{-}m}$ from the dataset $\mathcal{D}$ (described in Sec. \ref{sec:exp-datasets}).
Having both datasets allows us to compare the two different training strategies: training with a single target (quasi-optimal solution), and training with the best solutions found (which include the quasi-optimal solutions).
The data is divided into training, validation, and test sets.
The training sets contain the 200 small instances from $\mathcal{D}$, with 9, 13 or 18 jobs, while the validation and test sets contain, each, 20 large instances from $\mathcal{D}$, with 20, 22 or 24 jobs, such that no instance is used in both validation and test (empty intersection).

The models are trained according to the description provided in Sec. \ref{sec:meth-sol-pred}.
Both when training with the best solution for each instance and when training with multiple solutions, the models are trained using Adam to minimize the BCE between the prediction and the targets.

We tune the hyperparameters of the models by performing a random search and evaluating the models' performance on the validation set.
More specifically, we optimize the learning rate, number of convolutional layers ($L$), number of hidden features ($d$), graph convolutional operator, and whether or not the convolutions share the parameters.
Table~\ref{tab:hp-tuning} shows the ranges and best values found.
\new{%
We provide further details on the hyperparameter tuning, including a sensitivity analysis, in \ref{appx:hp-tuning}.  
Nonetheless, we underscore that the most impactful hyperparameters were the learning rate and the choice of graph convolution.
In particular, the choice of the SAGE convolutional operator was strongly correlated with lower validation losses for both the models trained with the best solution and those trained with multiple targets.
}

\begin{table}[h]
    \centering
    \caption{Hyperparameter space and best values for both models trained for optimality prediction. \emph{BS} and \emph{MS} indicate, resp., the models trained with the best solution and multiple solutions as targets. \emph{Weight tying} refers to the parameter sharing process between the two graph convolutional operators in the model (see Section \ref{sec:sat-gnn}).}
    \label{tab:hp-tuning}
    \begin{tabular}{l|c|c|c}
    \toprule
                             &                                               & \multicolumn{2}{c}{Best value} \\
    Hyperparameter           & Search space                                  & BS             & MS            \\
    \midrule
    Learning rate            & $\{10^{-2},10^{-3},10^{-4}\}$                 & $10^{-2}$      & $10^{-3}$     \\
    Layers ($L$)             & $\{1,2,3\}$                                   & $2$            & $3$           \\
    Hidden features ($d$)    & $\{2^5,2^6,2^7,2^8\}$                         & $2^6$          & $2^8$         \\
    Graph convolution        & $\{\text{GraphConv},\text{SAGE}\}$            & SAGE           & SAGE          \\
    Weight tying             & $\{\text{Yes},\text{No}\}$                    & Yes            & Yes           \\
    \bottomrule
    \end{tabular}
\end{table}

\begin{figure}
    \centering
    \begin{subfigure}{0.49\textwidth}
        \centering
        \includegraphics[width=\textwidth]{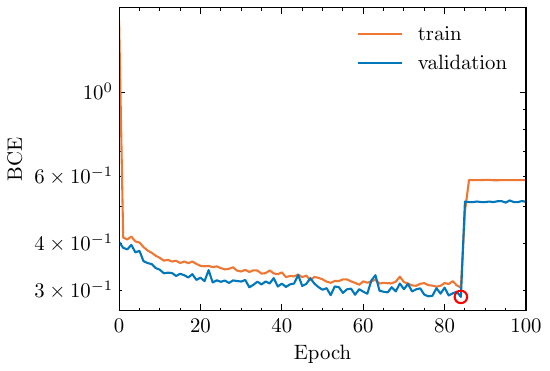}
        \caption{Best Solution}\label{fig:training-bs}
    \end{subfigure}
    \begin{subfigure}{0.49\textwidth}
        \centering
        \includegraphics[width=\textwidth]{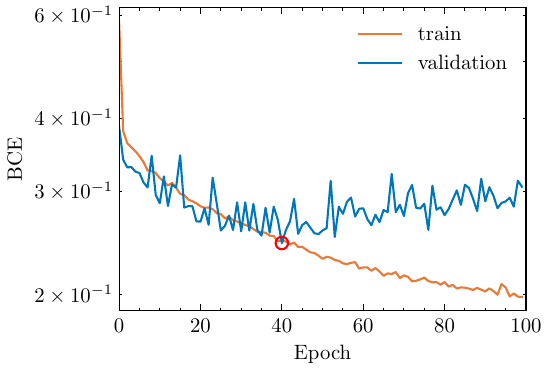}
        \caption{Multiple Solutions}\label{fig:training-ms}
    \end{subfigure}
    \caption{Training curves for SatGNN models trained with (a) the best solution or (b) multiple solutions. The best average BCE on the validation set (highlighted in red) is used for early-stopping the training.}
    \label{fig:opt-training-curves}
\end{figure}

Our training stopping criteria is determined through the validation set, i.e., from all epochs of our training budget (100 epochs), we pick the model that achieves the lowest validation loss.
The training curve for both training approaches can be seen in Figure \ref{fig:opt-training-curves}.
After training with the best solution, the average BCE on the validation set is 0.2887 and on the test set 0.2873.
After training with multiple solutions, the average BCE on the validation set is 0.2451 and on the test set 0.2482.
Although the BCE values are not comparable across training approaches, the small difference between the validation and the test sets for both approaches indicates no sign of overfitting.

\begin{figure}
    \centering
    \begin{subfigure}{0.49\textwidth}
        \centering
        \includegraphics{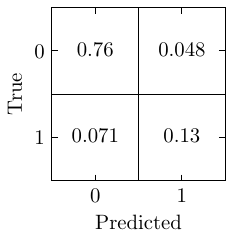}
        \caption{Best Solution}\label{fig:cm-bs}
    \end{subfigure}
    \begin{subfigure}{0.49\textwidth}
        \centering
        \includegraphics{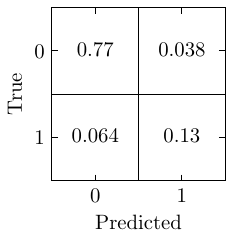}
        \caption{Multiple Solutions}\label{fig:cm-ms}
    \end{subfigure}
    \caption{Accuracy of the models when predicting optimal assignments. Computed with respect to all variables of all instances in the test set.}
    \label{fig:confusion-matrices}
\end{figure}

To grasp the actual performance of the models in the optimality prediction, we evaluate both on the test set after training.
Because of the high number of variables, neither could predict all variable assignments of any instance perfectly.
However, when considering each variable assignment individually, the model trained with multiple solutions and the model trained with the best solutions achieved 90\% and 88\% average accuracy, respectively.
We point out that the data is naturally biased towards zero assignments, as most jobs pass most of the time inactive (not assigned to run) when there are many jobs to be scheduled.
Given the imbalance of the data, the performance of both models is better visualized in Figure~\ref{fig:confusion-matrices}.
Furthermore, the model trained with multiple solutions achieved an F1 score of 0.72, while the model trained with the best solutions achieved an F1 score of 0.68.

\begin{figure}[h]
    \centering
    \includegraphics{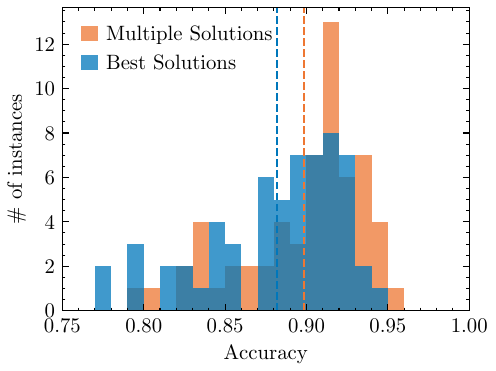}
    \caption{Histogram of accuracies for both models over the test set. For each instance, we compute the ratio of correctly predicted variable assignments. Vertical dashed lines indicate the mean value.}
    \label{fig:histogram-accuracy}
\end{figure}

For building useful primal heuristics, the models should achieve consistent behavior across instances, meaning they should correctly predict a high number of variable assignments.
Under that perspective, we measure the accuracy within each instance, which is calculated as the ratio of correctly classified variable assignments within a given instance.
The performance is detailed in the histogram of Figure~\ref{fig:histogram-accuracy}.
We see that both models achieve high expected accuracy values, and none of them ever correctly predicts less than 75\% of the variables correctly.

\begin{figure}[h]
    \centering
    \begin{subfigure}{0.49\textwidth}
        \centering
        \includegraphics[width=\textwidth]{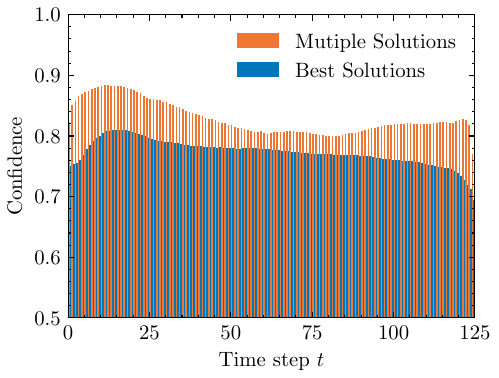}
        \caption{$x_{j,t}$}\label{fig:conf-x}
    \end{subfigure}
    \begin{subfigure}{0.49\textwidth}
        \centering
        \includegraphics[width=\textwidth]{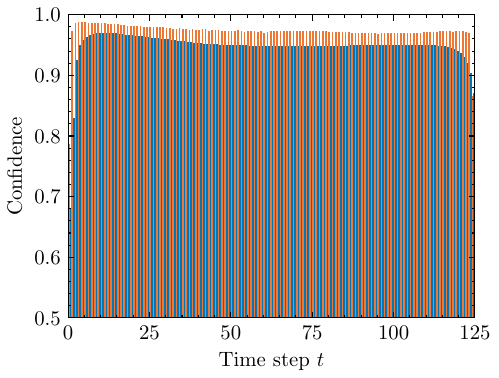}
        \caption{$\phi_{j,t}$}\label{fig:conf-phi}
    \end{subfigure}
    \caption{Average confidence (probability of the predicted class) of predicted values for (a) $\bm{x}$ and (b) $\bm{\phi}$ variables of the model trained only with the best solution and the model trained with multiple solutions. The average is taken over all jobs of all instances in the test set.}
    \label{fig:prediction-confidences}
\end{figure}

A deeper analysis is performed on the models' confidence when predicting variable assignments $\bm{\kappa}(\bm{z}=1|I)$, as presented in Sec. \ref{sec:meth-heuristics-ws}.
The average confidence of both models on instances of the test set is depicted in Figure \ref{fig:prediction-confidences} with respect to the binary variables of the optimization problem.
Training with multiple solutions seems to result in a more confident model overall.
Furthermore, both models provide significantly more confident predictions for the $\bm{\phi}$ variables.


\subsection{SatGNN-based Heuristic}\label{sec:exp-heuristics}

The two models presented in Sec. \ref{sec:exp-opt-pred} are used (each) to build three matheuristics, as described in Sec. \ref{sec:meth-sol-pred}.
Namely, the two models are the SatGNN trained with the best solution available and the SatGNN trained with multiple solutions (see Sec. \ref{sec:meth-sol-pred}).
The three matheuristics evaluated are: warm-starting, early-fixing, and trust region.
We use SCIP as our solver both for the baseline results and the optimization within the matheuristics.

Although all model parameters and hyperparameters are already defined (and will not change) by the time the heuristics are assembled, the heuristics have their own hyperparameters, which require adjustment.
More specifically, all three heuristic approaches have the hyperparameter $N\in\mathbb{N}$, representing the size of the partial solution extracted from SatGNN.
Besides, the trust region matheuristic has the size of the trust region $\Delta\in\mathbb{N}$.

We use the validation set to select the best values for the hyperparameters, aiming to optimize two performance metrics.
For once, we select values for the hyperparameters that maximize the expected objective value.
The objective values are normalized, i.e., we divide them by the objective value of the best solution known for each instance\footnote{We remind the reader that those solutions were computed along with the dataset generation, as detailed in Section~\ref{sec:exp-datasets}.}, resulting in a value between $0$ (no QoS) and $1$ (max. QoS known for the instance).
In parallel, we select the hyperparameters that minimize the time required to find a feasible solution to each problem.
The two tunings are performed for the SatGNN model trained with the best solution available and the SatGNN trained with multiple solutions.
Table \ref{tab:best-N-delta} summarizes the best hyperparameters for each model and each heuristic approach, in both evaluations.

\begin{table}[h]
    \centering
    \caption{Best values for partial solution size ($N$) and trust region size ($\Delta$) for each heuristic approach and each SatGNN model, as evaluated on the validation set. \emph{Objective} corresponds to the values that maximize the objective value of the best solution found. \emph{Feasibility} indicates the values that minimize the amount of time to find a feasible solution.}
    \label{tab:best-N-delta}
    \begin{tabular}{ll|cc|cc}
    \toprule
                                        &              & \multicolumn{2}{c|}{Objective} & \multicolumn{2}{c}{Feasibility} \\
    SatGNN model                        & Heuristic    & $N$         & $\Delta$        & $N$          & $\Delta$         \\
    \midrule
    \multirow{3}{*}{Best Solution}      & Warm-start   & 750         & -               & 1000         & -                \\
                                        & Early-fix    & 500         & -               & 750          & -                \\
                                        & Trust region & 1000        & 5               & 1000         & 1                \\
    \midrule
    \multirow{3}{*}{Multiple Solutions} & Warm-start   & 1750        & -               & 1500         & -                \\
                                        & Early-fix    & 1000        & -               & 1250         & -                \\
                                        & Trust region & 1250        & 1               & 1750         & 1             \\
    \bottomrule
    \end{tabular}
\end{table}

\begin{figure}
    \centering
    \begin{subfigure}{0.99\textwidth}
        \centering
        \includegraphics[width=\textwidth]{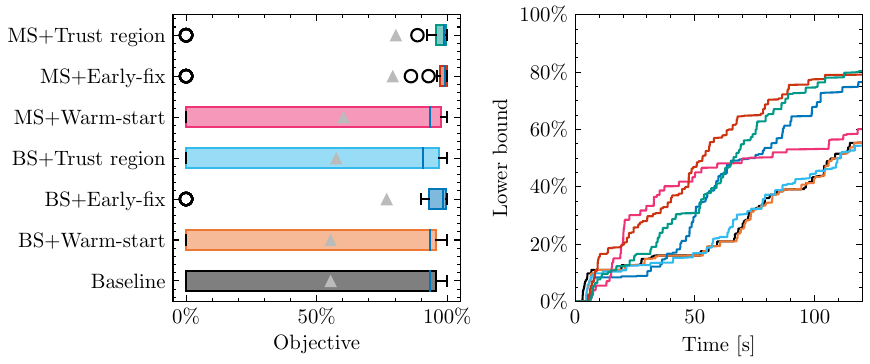}
        \caption{Best solution found within the time budget.}
        \label{fig:heuristics-test-results-obj}
    \end{subfigure}
    \begin{subfigure}{0.99\textwidth}
        \centering
        \includegraphics[width=\textwidth]{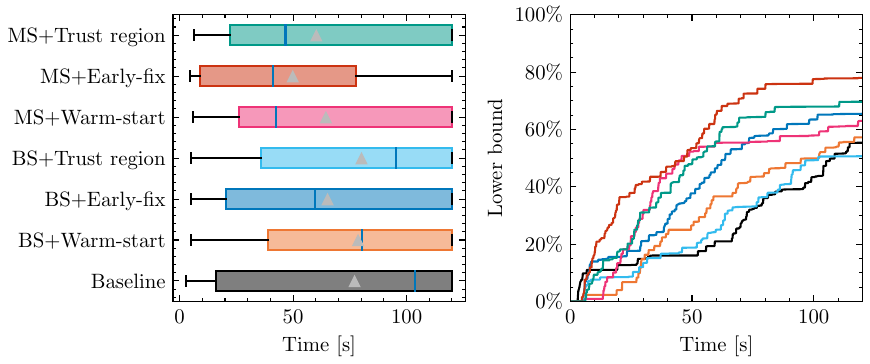}
        \caption{Time to find a feasible solution.}
        \label{fig:heuristics-test-results-feas}
    \end{subfigure}
    \caption{Test set performance of the SatGNN-based matheuristics. On the left, we have the distribution of the evaluation metric of interest over the instances of the test set for the multiple approaches, in which the triangle indicates the mean value and the circles indicate outliers. \emph{MS} indicates that the SatGNN model trained with multiple solutions was used, whereas \emph{BS} indicates that the model trained solely with the optimal solution was used instead. On the right is the average progress of the lower bound on all test set instances. The objective value is considered relative to the objective of the best solution known, thereby always lying in the unit interval. The heuristics' hyperparameters $N$ and $\Delta$ are defined upon experiments on the validation set, as presented in Table \ref{tab:best-N-delta}.}
    \label{fig:heuristics-test-results}
\end{figure}

The heuristics are then evaluated on the test set, which is not used for tuning neither the deep learning models' hyperparameters nor the heuristics' hyperparameters.
The final performance is measured through the expected relative objective value given the time budget, and the time it takes the heuristics to find a feasible solution.
Furthermore, we also analyze the progress of the lower bound over time, indicating the expected performance under more restricted budgets.
Figure \ref{fig:heuristics-test-results} illustrates the results.

\begin{figure}
    \centering
    \begin{subfigure}{0.99\textwidth}
        \centering
        \includegraphics{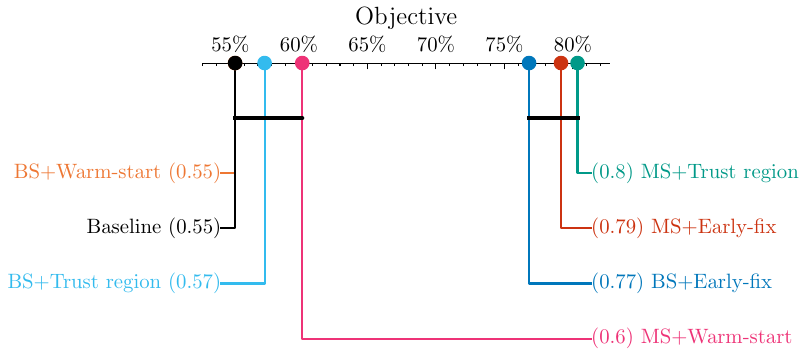}
        \caption{Best solution found within the time budget.}
        \label{fig:cdd-obj}
    \end{subfigure}
    \begin{subfigure}{0.99\textwidth}
        \centering
        \includegraphics{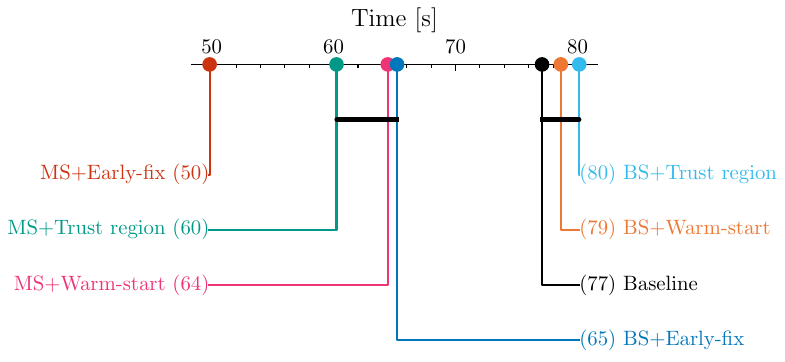}
        \caption{Time to find a feasible solution.}
        \label{fig:cdd-feas}
    \end{subfigure}
    \caption{Critical difference diagram presenting the average test set performance of the SatGNN-based matheuristics (round marker in the axis). A crossbar between two (or more) approaches indicates that their performance (distribution on the test set) was not deemed significantly different ($p$-value$>0.05$) through the paired Wilcoxon signed-rank test~\citep{wilcoxon_1945}.}
    \label{fig:cdds}
\end{figure}

To assess the significance of the results presented in Figure \ref{fig:heuristics-test-results} through statistical tests.
Given that the matheuristic performance data is generated from a shared set of instances, our samples are paired.
Furthermore, there is no evidence of normality in the results' distribution.
Therefore, we apply the Wilcoxon signed-rank test~\citep{wilcoxon_1945}, which is a non-parametric version of the t-test for matched pairs.
We apply the test pair-wise, comparing each matheuristic approach to every other matheuristic approach, and to the baseline.
The results are summarized in the critical difference diagram of Figure \ref{fig:cdds}.
Through early-fixing, both SatGNN models provide statistically significant ($p$-value$>0.05$) improvements over the baseline in both performance metrics.
However, the results show a significant advantage of the model trained with multiple solutions, providing better solutions (objective value) through early-fixing and trust region, and speeding up the feasibility definition (time to find a feasible solution) through all heuristic approaches.

Through the SatGNN model trained with multiple solutions, a 43\% increase in the expected objective value within the time budget is achieved by early-fixing the partial solution, while defining a trust region around that same partial solution resulted in an expected 45\% increase.
Although the results are close, we see from the progress of the lower bound (in Fig. \ref{fig:heuristics-test-results-obj}, the plot on the right) that the early-fix heuristic is able to find better solutions more quickly than the trust region method during the time budget.
In terms of feasibility, the early-fixing strategy using the same SatGNN model reduces in 35\% the expected time to find a feasible solution, having a significant advantage over all other heuristic approaches.
Surprisingly, even when optimized (through hyperparameter tuning) for reducing the time to find a feasible solution, the early-fix heuristic still improves the expected objective value of the best solution found in 41\%, as the lower bound progress illustrates (in Fig. \ref{fig:heuristics-test-results-feas}, the plot on the right).

\subsubsection{\new{Sensitivity Analysis}}\label{sec:exp-N}

\begin{figure}
    \centering
    \begin{subfigure}[t]{0.59\textwidth}
        \includegraphics[width=\textwidth]{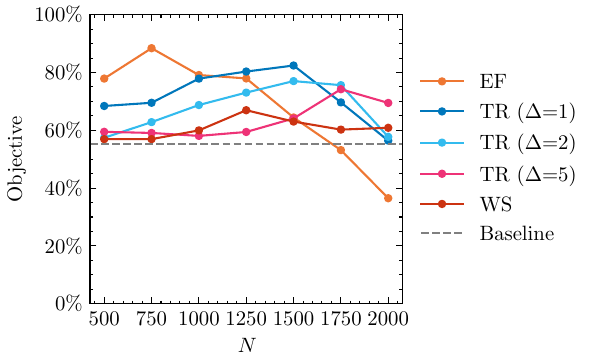}
    \end{subfigure}
    \begin{subfigure}[t]{0.40\textwidth}
        \includegraphics[width=\textwidth]{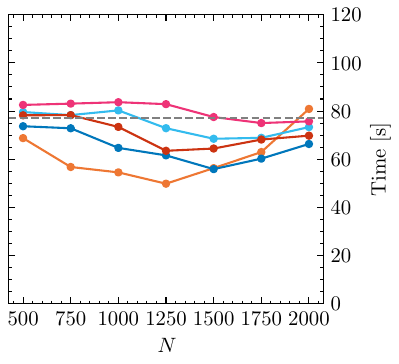}
    \end{subfigure}
    \caption{\new{Sensitivity analysis of the solution methods with respect to the size of the partial solution $N$.} \emph{EF} indicates the early-fixing matheuristic, \emph{TR} the trust region, and \emph{WS} warm-start. The SatGNN model trained with multiple solutions is used. The evaluation is performed on the test set.}
    \label{fig:N-impact}
\end{figure}

\new{%
The proposed solution methods share a common parameter: the number $N$ of variables in the partial solution.
To evaluate the sensitivity of the results to this parameter, we evaluate the SatGNN model on the test set for varying values of $N$.
}
More specifically, we vary the value of $N$ for all matheuristics and measure the impact on the test set considering both perspectives (objective value and time to find a feasible solution).
For conciseness, we consider only the SatGNN model trained with multiple solutions, as it is overall superior, in our experiments, to the model trained solely with the best solution.
The results are illustrated in Figure \ref{fig:N-impact}.
 
\new{%
Beyond $N$,  the trust region method has an additional parameter: the size of the trust region, $\Delta$.
To evaluate the full potential of this matheuristic, we also consider three different values for $\Delta$ in our sensitivity analysis experiments.
These results are also displayed in Figure~\ref{fig:N-impact}, as the legends indicate.
}

\begin{figure}
    \centering
    \includegraphics{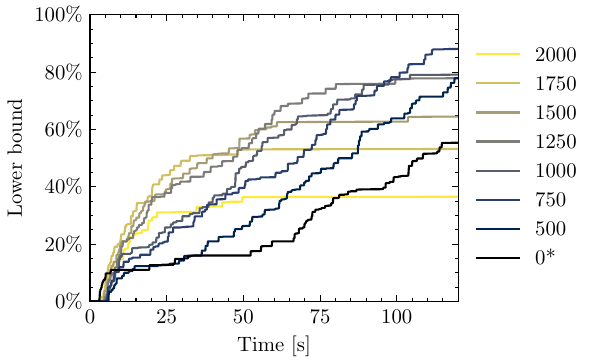}
    \caption{\new{Average progress of the lower bound of SatGNN-based early fixing on the test set for varying values of the partial solution size $N$.}}
    \label{fig:N-impact-ef}
\end{figure}

As expected, the number of variables that compose the partial solution is impactful for all approaches, with early-fixing being the one most sensitive to it, and warm-start the less sensitive one.
\new{%
In fact, analyzing the results for the early-fixing approach in detail, we note that $N$ indeed imposes a trade-off between solution quality and speed, as shown in Figure~\ref{fig:N-impact-ef}, confirming the intuitive interpretation of this parameter discussed in Section~\ref{sec:meth-heuristics-ef}.
More specifically, increasing the partial solution size seems to speed up the solver, allowing it to find better solutions more quickly, while reducing the final solution quality at the end of the time budget.
}

Furthermore, we note a relation between $N$ and $\Delta$ through the peak performance of the trust region method (including early-fixing, which is equivalent to trust region with $\Delta=0$).
In other words, larger partial solutions seem to require larger trust regions to achieve peak performance.
Intuitively, this result is expected since the partial solutions are based on confidence, and, thus, the larger the partial solution is, the higher the expected probability of it including wrongly predicted variables.
In turn, it suggests that the SatGNN model is properly trained (its confidence correlates to its performance).

\section{Discussion}\label{sec:discussion}

In this section, we discuss the results obtained from a series of experiments to address the research questions posed in the introduction.
The experiments aimed to evaluate the effectiveness of GNNs in solving the ONTS problem.
More specifically, we use our proposed model, SatGNN, based on the models by \citet{gasse_exact_2019} and \citet{hamilton_inductive_2017}.
Our experiments evaluate whether SatGNN can learn feasibility and optimality given large instances of the ONTS problem and whether the trained model can be used in heuristic solutions.

The results of the first set of experiments (Sec. \ref{sec:exp-feas}) indicate that SatGNN can provide accurate feasibility estimations even in challenging scenarios.
Three scenarios of increasing difficulty were considered, and the results were quite promising, with over 89\% accuracy in all experiments.
In the most challenging scenario, SatGNN is evaluated on large ($J\in\{20, 22, 24\}$) unseen instances, but trained on small instances ($J\in\{9,13,18\}$), and it achieved 94.15\% accuracy in feasibility classification.
The remarkable accuracy indicates that SatGNN is highly effective at learning the constraints of the ONTS problem.
This performance is expected as the feasibility classification task requires the model to learn linear classification boundaries, which are derived from the problem's linear constraints.
In other words, the model learns the indicator function of a convex set (considering the problem's LP relaxation), for which neural networks are known to be suitable.

The second set of experiments (Sec. \ref{sec:exp-opt-pred}) aimed to predict optimal solutions to ONTS instances using SatGNN.
More specifically, SatGNN was trained to predict the biases of the binary variables for multiple instances of the ONTS problem.
Two training approaches were explored: one trained with a single target (the quasi-optimal solution) and another trained with multiple solutions (the best feasible solutions found).
Both approaches showed promising results, with no signs of overfitting, as indicated by the small difference between validation and test set performance.

Both models achieved similar accuracy results on a per-variable basis, with the model trained with multiple solutions having a slight advantage of 2 p.p. on average.
Nevertheless, as illustrated by Figure~\ref{fig:histogram-accuracy}, both models show similar distributions, always correctly predicting more than 75\% of the variables in all instances.
Interestingly, the model trained with multiple solutions tended to be more confident in its predictions, particularly for the binary variables.
This increased confidence could be valuable in guiding the optimization process, as it can be understood as an indication of the model's certainty about its predictions.

Although the results from the second set of experiments suggest that SatGNN can effectively predict optimal or near-optimal solutions to ONTS instances, we put the models to the test through a third set of experiments (Sec. \ref{sec:exp-heuristics}) in which we build (mat)heuristic solutions based on the output of the SatGNN trained for optimality prediction.
Namely, as detailed in Sec. \ref{sec:meth-heuristics}, three approaches were evaluated for applying the model's output: warm-start, early fix, and trust region.
Each approach is optimized for finding the best solution, giving the time budget (2 min.) and reducing the time to find a feasible solution (2 separate experiments).
First, we notice that in all direct comparisons, the SatGNN model trained with multiple solutions resulted in better heuristic solutions than the model trained with a single solution, as the results of the second set of experiments suggested.

Early-fixing the partial solution provided by the SatGNN models yielded the best results overall, showing a 43\% gain in the evaluation of the expected objective value within the time budget, and a 35\% reduction in the evaluation of the expected time to find a feasible solution.
The trust region approach, which can be seen as a relaxation of early-fixing, was comparable to early-fixing in terms of the best solution found within the time limit. However, overall, it seems to provide a worse trade-off, as both early-fixing and trust region do not provide optimality guarantees, but trust region takes longer to find good solutions to the ONTS problem.
Using the SatGNN models to warm-start the SCIP solver significantly reduced the time needed to find a feasible solution (in comparison to the baseline), but it was worse overall than the two other approaches.

Across all experiments, a highlight of the GNNs' performance was the generalization to larger instances.
This characteristic is essential for problems with varying instance size and is not trivially achievable with traditional deep-learning models.
However, it is also key for tasks with higher acquisition costs for certain problem sizes, as in the case of the ONTS problem.
Large instances of optimization problems are the ones that drive the research of heuristic solutions, as they are usually more expensive to solve; however, as we need to solve instances to create a training dataset, it often becomes infeasible to train the deep learning models on large instances.
In this context, the generalization results from our experiments suggest that GNNs can be effectively trained using easy (cheap to acquire) instances of the problem, and be used on the hard instances of interest.

Generating reliable instances to train and evaluate solvers is a key point in such a project, as the confidence of the results depends on how carefully instances are designed to compose the test set~\citep{smith-miles_generating_2015}.
Data generation is particularly critical when historical data is not available~\citep{bengio_machine_2021}, which motivates the research on algorithms for generating reliable instances~\citep{smith-miles_generating_2015,malitsky_structure-preserving_2016}.
We follow the problem definition of the FloripaSat-I mission~\citep{marcelino_critical_2020} to define the parameters' ranges used to sample new instances uniformly.
Although the data generation process is reliable, it still depends on the optimization of every randomly generated instance.
Furthermore, as we restrict ourselves to feasible instances, we only add to the dataset instances for which the solver could find at least one feasible solution within the time budget, effectively restricting our data to an easier sub-problem~\citep{yehuda_its_2020}.

\new{%
Finally, we underscore that the computational gains observed in terms of time to find a feasible solution (see, for example, Figure~\ref{fig:cdd-feas}) are expected to directly translate to the deployed application.
First, because dataset generation and training, which are indeed costly, as discussed above, are performed offline.
  Then, generating a partial solution with SatGNN takes a negligible amount of time compared to solving the instances.  
  Beyond that, the inference time scales linearly with the size of the graph, which is directly proportional to the instance size (number of variables and constraints).
Given that the problem is combinatorial, we can expect the SatGNN inference cost to be totally dominated by the cost of solving the instances for realistic instances.
}

\section{Conclusion and Future Research}\label{sec:conclusion}

This work has proposed a novel approach to tackle the ONTS problem using graph neural networks.
More specifically, we investigated whether a GNN can learn the structure of the problem and be used to build effective heuristic solutions.
Our experiments showed that our proposed architecture, SatGNN, successfully classifies the feasibility of candidate solutions and is also able to provide reliable partial solutions to varied instances of the ONTS problem.
Not only was the model able to generalize to unseen instances, but it also showed promising results on out-of-distribution instances, which were larger than the ones seen during training.
This outcome shows how the inherent symmetries of graph neural networks make them suitable for dealing with the structures of the optimization problem.

By leveraging partial solutions provided by the SatGNN model, we built heuristic solutions to the ONTS problem that outperformed the SCIP solver in hard instances.
More specifically, by fixing the variable assignment from the partial solution provided by SatGNN, our heuristics were able to improve up to 43\% in the expected objective value of the best solution found within the time budget and reduce up to 35\% in the expected time to find a feasible solution.

Overall, the experiments presented in this paper showcase the potential of GNNs for addressing the ONTS problem.
The high accuracy in feasibility classification, the ability to predict optimal solutions, and the effectiveness of SatGNN-based heuristics suggest that our learning-based approach yields a strong primal heuristic that has the potential to significantly improve the performance of exact methods for complex space mission scheduling problems.


\subsection{Future Work}

Given the nature of the ONTS problem, generating large, difficult instances is a significant challenge.
Although realistic, the instance generation method used in this work is limited regarding instance difficulty by the time budget.
One possible future research direction is to leverage easier instances, combining job parameters to generate large feasible instances that extrapolate the time budget restriction.
This line also leads to a non-trivial application of machine learning models for combinatorial optimization: guiding the search for difficult instances.
More specifically, using the SatGNN for feasibility classification could significantly alleviate the cost of generating larger feasible instances.

\new{%
Reinforcement learning (RL) has shown promise for learning-based approaches to combinatorial optimization~\citep{jalalikhalilabadiDeepReinforcementLearningbased2024,omalleyReinforcementLearningMixedinteger2023,waubertdepuiseauReliabilityReinforcementLearning2022}.
It has also been explored in settings close to ONTS, such as Earth observation satellite scheduling~\citep{heGenericMarkovDecision2022,jacquetEarthObservationSatellite2024}.
The possibility of taking sequential decisions makes it easier to incorporate feasibility information into the model.
In our case, it could enable the fixing of larger partial solutions without rendering the resulting mathematical program infeasible, reducing the computational cost.
It may also reduce dataset-generation effort by learning from interaction rather than labeled optimal solutions.
However, developing an RL-based primal heuristic requires careful reward design and an interactive environment, which is substantially different from our supervised learning approach; we therefore leave a full RL solution to future work.
}

Future works naturally entail the integration of the proposed methodology into exact methods tailored to the ONTS problem.
State-of-the-art branch-and-price algorithms~\citep{rigo_branch-and-price_2022} could significantly benefit from better initial solutions.
The explainability aspects could be further developed to provide deeper insights into solution quality and guide the generation of columns for specific jobs.
Additionally, the approach could be adapted to handle dynamic scheduling scenarios where task requirements evolve over time, requiring online learning to update the parameters of the SatGNN.
Also, the principles developed here could be extended to related space mission optimization problems, including constellation management and multi-satellite resource allocation challenges.

\section*{Acknowledgments}

The authors acknowledge support from CNPq (\emph{Conselho Nacional de Desenvolvimento Cient{\'\i}fico e Tecnol{\'o}gico}) under grant numbers 150281/2022-6 and 404576/2021-4 as well as FAPESC under grant number 2021TR001851.

\section*{Declaration of Generative AI and AI-assisted technologies in the writing process}

During the preparation of this work the author(s) used generative AI tools in order to perform light editing of the text. After using these tools/services, the author(s) reviewed and edited the content as needed and take(s) full responsibility for the content of the publication.

\appendix

\section{Symbols in the ONTS mathematical programming model}\label{appx:tab-symbols}

The table below compiles the sets, variables, and parameters of the ONTS MILP formulation.

\bigskip

\begin{tabular}{cl}
\hline
\textbf{Symbol} & \textbf{Description} \\ \hline
\multicolumn{2}{l}{\quad\textit{Sets}} \\ \hline
\(\mathcal{J}\) & Tasks assigned to the nanosatellite \(\{1, \dots, J\}\) \\ 
\(\mathcal{T}\) & Discrete time periods within the planning horizon \(\{1, \dots, T\}\) \\ \hline
\multicolumn{2}{l}{\quad\textit{Decision Variables}} \\ \hline
\(x_{j,t}\) & Binary: 1 if task \(j\) is executed at time \(t\), 0 otherwise \\
\(\phi_{j,t}\) & Binary: 1 if task \(j\) starts at time \(t\), 0 otherwise \\
\hline
\multicolumn{2}{l}{\quad\textit{Parameters}} \\ \hline
\(u_{j,t}\) & Utility or mission value derived from executing task \(j\) at time \(t\) \\ 
\(w^{\min}_{j}\) & Minimum start time window for task \(j\) \\
\(w^{\max}_{j}\) & Maximum start time window for task \(j\) \\ 
\({p}^{\min}_{j}\) & Minimum time between consecutive starts of task \(j\) \\ 
\({p}^{\max}_{j}\) & Maximum time between consecutive starts of task \(j\) \\ 
\({t}^{\min}_{j}\) & Minimum execution time of task \(j\) \\ 
\({t}^{\max}_{j}\) & Maximum execution time of task \(j\) \\ 
\({y}^{\min}_{j}\) & Minimum number of start-ups for task \(j\) \\ 
\({y}^{\max}_{j}\) & Maximum number of start-ups for task \(j\) \\ 
\(b_{t}\) & Energy balance at time \(t\) \\ 
\(i_{t}\) & Total current at time \(t\) \\ 
\(\text{SoC}_{t}\) & State-of-charge of the battery at time \(t\) \\  
\(r_{t}\) & Power input at time \(t\) (from solar panels) \\  
\(q_{j}\) & Power consumption of task \(j\) per time slot \\ 
\(V_{b}\) & Nominal battery voltage \\  
\(e\) & Battery efficiency during charging and discharging \\  
\(Q\) & Nominal capacity of the battery \\ 
\(\rho\) & Minimum energy level allowed for the battery \\  
\(\gamma\) & Maximum allowed current rate for charging or discharging the battery \\ \hline
\end{tabular}
\color{black}

\section{Random instance generation}\label{appx:random-instance}

For any particular mission size and orbital length, the main objective is to generate a realistic ONTS case using random data.
We have taken the FloripaSat-I mission as a reference for instance generation, which has an altitude of 628 kilometers and an orbital period of 97.2 minutes~\cite{marcelino_critical_2020}.
The battery-related parameters are fixed for all instances as
\begin{align*}
    e &\gets 0.9 \\
    Q &\gets 5 \\
    \gamma &\gets 5 \\
    V_b &\gets 3.6 \\
    \rho &\gets 0.0
\end{align*}
The attitude considered here is the Nadir, in which the satellite turns at the same rate around the Earth, so one side (or axis) always faces the Earth's surface.
This analytical model then utilizes a rotation matrix to simulate the satellite's dynamics and can be adapted for larger or different geometries by adjusting the normal vectors representing the body.

To generate realistic power input vectors $\boldsymbol{r}$, we use 2 years of historical data of solar irradiance in orbit.
More precisely, a window is randomly drawn from the historical data, and an analytical model is used to determine the power input vector.
Once orbits are stable and solar flux constant -- $1360 W/m^2$  -- one can calculate this vector by knowing the spacecraft orbit, attitude -- its kinematics -- and size~\cite{filho_comprehensive_2020}.

The remaining parameters are drawn uniformly, given handcrafted limits to increase the feasibility rate.
Specifically, given the number of tasks $J$ and the number of time units $T$, the parameters are drawn as
\begin{align*}
    u_j &\gets \mathcal{U}(1, J) \\
    q_j &\gets \mathcal{U}(0.3, 2.5) \\
    y_j^{\min} &\gets \mathcal{U}(1, \lceil T/45 \rceil) \\
    y_j^{\max} &\gets \mathcal{U}(y_j^{\min}, \lceil T/15 \rceil) \\
    t_j^{\min} &\gets \mathcal{U}(1, \lceil T/10 \rceil) \\
    t_j^{\max} &\gets \mathcal{U}(t_j^{\min}, \lceil T/4 \rceil) \\
    p_j^{\min} &\gets \mathcal{U}(t_j^{\min}, \lceil T/4 \rceil) \\
    p_j^{\max} &\gets \mathcal{U}(p_j^{\min}, T) \\
    w_j^{\min} &\gets \mathcal{U}(0, \lceil T/5 \rceil) \\
    w_j^{\max} &\gets \mathcal{U}(\lfloor T-\lceil T/5 \rceil \rfloor, T)
.\end{align*}

\section{\new{Symbols for SatGNN}}
\label{appx:gnn-symbols}

\new{The table below provides a summary of the main symbols used in the definition of SatGNN and the related methods.}

\begin{tabular}{cl}
\hline
\textbf{Symbol} & \textbf{Description} \\



\hline
\multicolumn{2}{l}{\quad\textit{Graph representation}} \\ \hline
$V_{\text{var}}$ & Nodes associated to variables; $|V_{\text{var}}|=n$. \\
$V_{\text{con}}$ & Nodes associated to constraints; $|V_{\text{con}}|=m$. \\
$E$ & Edges between variable nodes and constraint nodes. \\
$w(e)$ & Weight of edge $e\in E$. \\
$\mathcal{N}(v)$ & Neighborhood of node $v\in V$. \\

\hline
\multicolumn{2}{l}{\quad\textit{Features and hidden representations}} \\ \hline
$d_\text{in}$ & Number of input features for each node. \\
$\bm{f}^{v}$ & Input feature vector of node $v$. \\
$F_{\text{var}}$ & Matrix of hidden features for variable nodes. \\
$F_{\text{con}}$ & Matrix of input features for constraint nodes. \\
$L\in\mathbb{N}$ & Number of graph convolution layers. \\
$d\in\mathbb{N}$ & Number of hidden features for each node. \\
$\bm{h}^{(l)}_{v}$ & Hidden feature vector of node $v$ at layer $l\in\{0,\dots,L\}$. \\
$H^{(l)}_{\text{var}}$ & Matrix of hidden features for variable nodes at layer $l$. \\
$H^{(l)}_{\text{con}}$ & Matrix of hidden features for constraint nodes at layer $l$. \\
$d_\text{out}$ & Number of output features for each node. \\


\hline
\multicolumn{2}{l}{\quad\textit{Fully-connected modules}} \\ \hline
$\mathrm{NN}_{\text{var}}$ & Encoder for variable-node features (produces $\bm{h}^{(0)}$ on $V_{\text{var}}$). \\
$\mathrm{NN}_{\text{con}}$ & Encoder for constraint-node features (produces $\bm{h}^{(0)}$ on $V_{\text{con}}$). \\
$\mathrm{NN}_{\text{out}}$ & Output head applied to $\bm{h}^{(L)}$ (task-dependent output dimension). \\



\hline
\multicolumn{2}{l}{\quad\textit{Training}} \\ \hline
$\mathcal{L}$ & Loss function. \\
$\mathcal{D}$ & Dataset of instances and candidate solutions. \\
$p(\hat{\bm{z}} \in Z | I)$ & True ``probability'' that the assignment $\hat{\bm{z}}$ is feasible for instance $I$. \\ & Because there is no stochasticity, it can either be 0 or 1. \\
$\hat{p}(\hat{\bm{z}} \in Z | I)$ & Estimated ``probability'' (predicted variable bias) that the assignment \\ & $\hat{\bm{z}}$ is feasible for instance $I$. \\
$\hat{p}(\bm{z}^* = 1 | I)$ & Predicted variable biases (towards 1) of an optimal solution for \\ & instance $I$. \\

\hline
\multicolumn{2}{l}{\quad\textit{Heuristics}} \\ \hline
$\kappa(z|I)$ & Confidence the model has in each variable bias prediction. Defined as \\ & the maximum over the predicted marginal probabilities. \\
$\mathcal{C}^N_I$ & set of indices of the $N\in \mathbb{N}$ variables that the model is most confident \\ & with with respect to its prediction for instance $I$. \\
$\Delta$ & Size of the trust region around a predicted partial assignment. \\

\hline
\end{tabular}
\color{black}

\section{\new{Hyperparameter Tuning}}
\label{appx:hp-tuning}


\new{%
We conduct a random search over the hyperparameters listed in Table~\ref{tab:hp-tuning} and evaluate each configuration on the validation set.
For each configuration we train SatGNN using the best solution (BS) and the multiple solution (MS) approaches on the training set.
The performance is measured as the best validation loss observed in 10 epochs.
The best configurations seen for each training method is reported in Table~\ref{tab:hp-tuning}.
}

\begin{figure}
    \centering
    \begin{subfigure}{\textwidth}
        \centering
        \includegraphics[width=\textwidth]{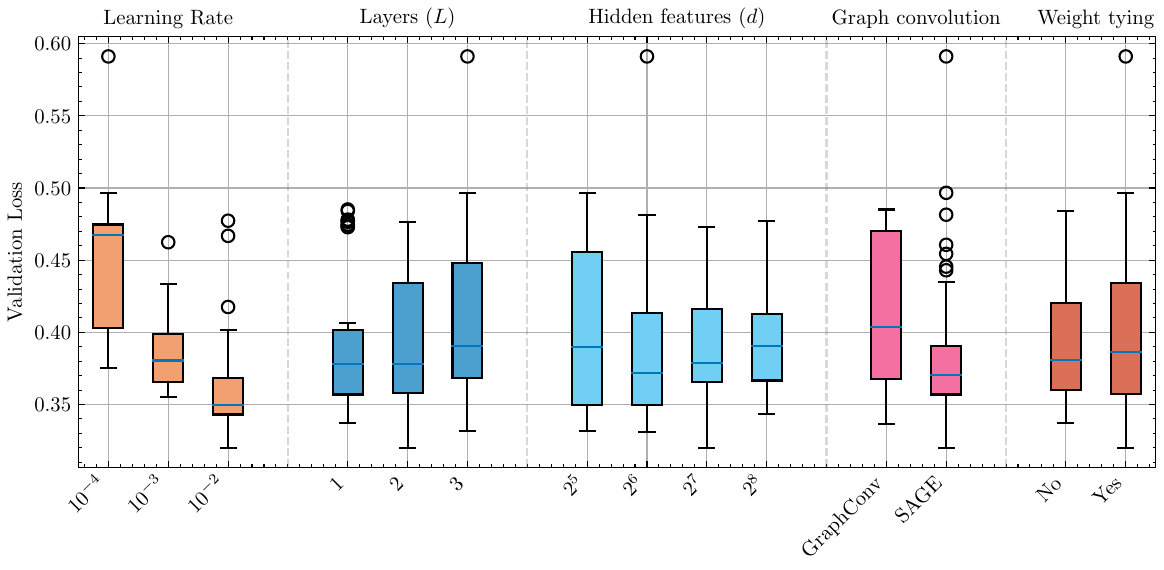}
        \caption{Best Solution}\label{fig:hp-tuning-bs}
    \end{subfigure}
    \begin{subfigure}{\textwidth}
        \centering
        \includegraphics[width=\textwidth]{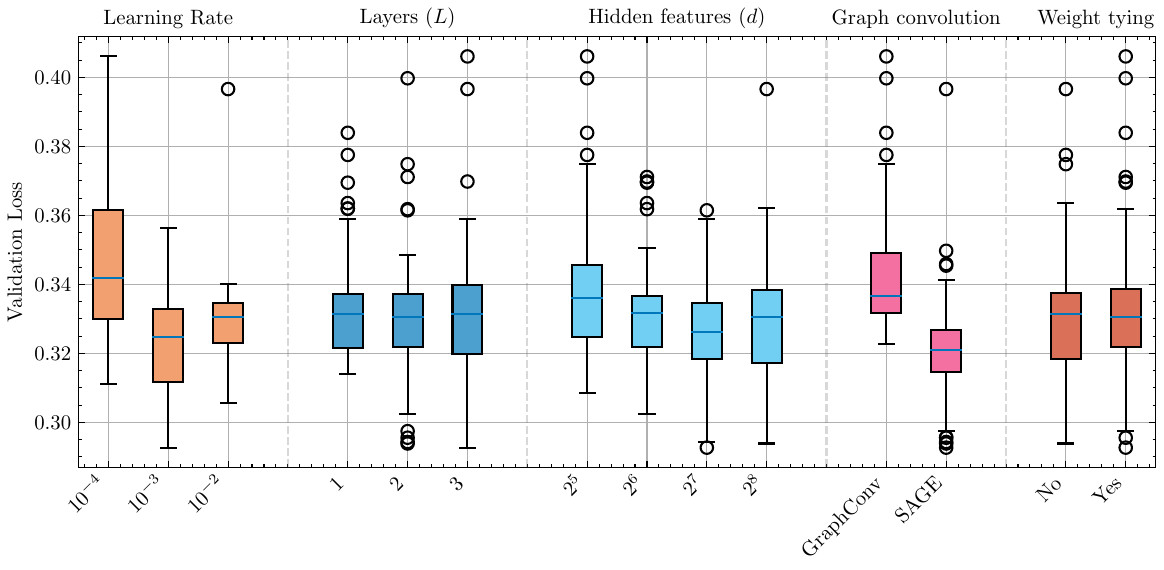}
        \caption{Multiple Solutions}\label{fig:hp-tuning-ms}
    \end{subfigure}
    \caption{Validation loss by hyperparameter choice in our hyperparameter tuning experiments.}
    \label{fig:hp-box}
\end{figure}

\new{%
We evaluate the relevance of tuning each hyperparameter by measuring how sensitive the validation loss of SatGNN is to the different values in search space.
The distribution of the validation loss given each hyperparameter choice is illustrated in Figure~\ref{fig:hp-box}.
For each hyperparameter, we test whether validation losses differ across its range using the nonparametric Kruskal–Wallis $H$ test~\citep{macfarlandKruskalWallisHTest2016}.
We apply the test independently for the BS and the MS trainings.
The only hyperparameters that had a statistically significant ($p \ll 0.05$) impact in the validation loss were the learning rate and the graph convolutional operator, for both training approaches.
All other hyperparameters did not achieve significance under this criterion, although the size of the hidden feature vectors $d$ had a strong impact ($H = 6.84$ and $p<0.08$) when training with MS.
}

\new{%
We refer the reader to our code repository\footnote{\url{https://github.com/brunompacheco/sat-gnn}} for further details, as we have made all our hyperparameter tuning implementation and results available.
}

\bibliographystyle{elsarticle-harv} 
\bibliography{compiled_main.bib}

\begin{thebibliography}{60}
\expandafter\ifx\csname natexlab\endcsname\relax\def\natexlab#1{#1}\fi
\providecommand{\url}[1]{\texttt{#1}}
\providecommand{\href}[2]{#2}
\providecommand{\path}[1]{#1}
\providecommand{\DOIprefix}{doi:}
\providecommand{\ArXivprefix}{arXiv:}
\providecommand{\URLprefix}{URL: }
\providecommand{\Pubmedprefix}{pmid:}
\providecommand{\doi}[1]{\href{http://dx.doi.org/#1}{\path{#1}}}
\providecommand{\Pubmed}[1]{\href{pmid:#1}{\path{#1}}}
\providecommand{\bibinfo}[2]{#2}
\ifx\xfnm\relax \def\xfnm[#1]{\unskip,\space#1}\fi
\bibitem[{Bengio et~al.(2021)Bengio, Lodi and Prouvost}]{bengio_machine_2021}
\bibinfo{author}{Bengio, Y.}, \bibinfo{author}{Lodi, A.},
  \bibinfo{author}{Prouvost, A.}, \bibinfo{year}{2021}.
\newblock \bibinfo{title}{Machine learning for combinatorial optimization: {A}
  methodological tour d’horizon}.
\newblock \bibinfo{journal}{European Journal of Operational Research}
  \bibinfo{volume}{290}, \bibinfo{pages}{405--421}.
\newblock \DOIprefix\doi{10.1016/j.ejor.2020.07.063}.
\bibitem[{Bestuzheva et~al.(2021)Bestuzheva, Besançon, Chen, Chmiela,
  Donkiewicz, Doornmalen, Eifler, Gaul, Gamrath, Gleixner, Gottwald, Graczyk,
  Halbig, Hoen, Hojny, Hulst, Koch, Lübbecke, Maher, Matter, Mühmer, Müller,
  Pfetsch, Rehfeldt, Schlein, Schlösser, Serrano, Shinano, Sofranac, Turner,
  Vigerske, Wegscheider, Wellner, Weninger and Witzig}]{bestuzheva_scip_2021}
\bibinfo{author}{Bestuzheva, K.}, \bibinfo{author}{Besançon, M.},
  \bibinfo{author}{Chen, W.K.}, \bibinfo{author}{Chmiela, A.},
  \bibinfo{author}{Donkiewicz, T.}, \bibinfo{author}{Doornmalen, J.v.},
  \bibinfo{author}{Eifler, L.}, \bibinfo{author}{Gaul, O.},
  \bibinfo{author}{Gamrath, G.}, \bibinfo{author}{Gleixner, A.},
  \bibinfo{author}{Gottwald, L.}, \bibinfo{author}{Graczyk, C.},
  \bibinfo{author}{Halbig, K.}, \bibinfo{author}{Hoen, A.},
  \bibinfo{author}{Hojny, C.}, \bibinfo{author}{Hulst, R.v.d.},
  \bibinfo{author}{Koch, T.}, \bibinfo{author}{Lübbecke, M.},
  \bibinfo{author}{Maher, S.J.}, \bibinfo{author}{Matter, F.},
  \bibinfo{author}{Mühmer, E.}, \bibinfo{author}{Müller, B.},
  \bibinfo{author}{Pfetsch, M.E.}, \bibinfo{author}{Rehfeldt, D.},
  \bibinfo{author}{Schlein, S.}, \bibinfo{author}{Schlösser, F.},
  \bibinfo{author}{Serrano, F.}, \bibinfo{author}{Shinano, Y.},
  \bibinfo{author}{Sofranac, B.}, \bibinfo{author}{Turner, M.},
  \bibinfo{author}{Vigerske, S.}, \bibinfo{author}{Wegscheider, F.},
  \bibinfo{author}{Wellner, P.}, \bibinfo{author}{Weninger, D.},
  \bibinfo{author}{Witzig, J.}, \bibinfo{year}{2021}.
\newblock \bibinfo{title}{The {SCIP} {Optimization} {Suite} 8.0}.
\newblock \bibinfo{type}{{ZIB}-{Report}} \bibinfo{number}{21-41}. Zuse
  Institute Berlin.
\newblock \URLprefix \url{http://nbn-resolving.de/urn:nbn:de:0297-zib-85309}.
\bibitem[{Boschetti and Maniezzo(2022)}]{boschetti_matheuristics_2022}
\bibinfo{author}{Boschetti, M.A.}, \bibinfo{author}{Maniezzo, V.},
  \bibinfo{year}{2022}.
\newblock \bibinfo{title}{Matheuristics: using mathematics for heuristic
  design}.
\newblock \bibinfo{journal}{4OR} \bibinfo{volume}{20},
  \bibinfo{pages}{173--208}.
\newblock \DOIprefix\doi{10.1007/s10288-022-00510-8}.
\bibitem[{Camponogara et~al.(2022)Camponogara, Seman, Rigo, Filho, Ribeiro and
  Bezerra}]{camponogara_continuous-time_2022}
\bibinfo{author}{Camponogara, E.}, \bibinfo{author}{Seman, L.O.},
  \bibinfo{author}{Rigo, C.A.}, \bibinfo{author}{Filho, E.M.},
  \bibinfo{author}{Ribeiro, B.F.}, \bibinfo{author}{Bezerra, E.A.},
  \bibinfo{year}{2022}.
\newblock \bibinfo{title}{A continuous-time formulation for optimal task
  scheduling and quality-of-service assurance in nanosatellites}.
\newblock \bibinfo{journal}{Computers \& Operations Research}
  \bibinfo{volume}{147}, \bibinfo{pages}{105945}.
\newblock \DOIprefix\doi{10.1016/j.cor.2022.105945}.
\bibitem[{Cappart et~al.(2022)Cappart, Chételat, Khalil, Lodi, Morris and
  Veličković}]{cappart_combinatorial_2022}
\bibinfo{author}{Cappart, Q.}, \bibinfo{author}{Chételat, D.},
  \bibinfo{author}{Khalil, E.}, \bibinfo{author}{Lodi, A.},
  \bibinfo{author}{Morris, C.}, \bibinfo{author}{Veličković, P.},
  \bibinfo{year}{2022}.
\newblock \bibinfo{title}{Combinatorial optimization and reasoning with graph
  neural networks}.
\newblock \DOIprefix\doi{10.48550/arXiv.2102.09544}.
\bibitem[{Cui et~al.(2023)Cui, Song, Zhang, Tao, Liu and Shi}]{9998480}
\bibinfo{author}{Cui, K.}, \bibinfo{author}{Song, J.}, \bibinfo{author}{Zhang,
  L.}, \bibinfo{author}{Tao, Y.}, \bibinfo{author}{Liu, W.},
  \bibinfo{author}{Shi, D.}, \bibinfo{year}{2023}.
\newblock \bibinfo{title}{Event-triggered deep reinforcement learning for
  dynamic task scheduling in multisatellite resource allocation}.
\newblock \bibinfo{journal}{IEEE Transactions on Aerospace and Electronic
  Systems} \bibinfo{volume}{59}, \bibinfo{pages}{3766--3777}.
\newblock \DOIprefix\doi{10.1109/TAES.2022.3231239}.
\bibitem[{Dalin et~al.(2021)Dalin, Haijiao, Zhen, Yanfeng and Shi}]{9152114}
\bibinfo{author}{Dalin, L.}, \bibinfo{author}{Haijiao, W.},
  \bibinfo{author}{Zhen, Y.}, \bibinfo{author}{Yanfeng, G.},
  \bibinfo{author}{Shi, S.}, \bibinfo{year}{2021}.
\newblock \bibinfo{title}{An online distributed satellite cooperative
  observation scheduling algorithm based on multiagent deep reinforcement
  learning}.
\newblock \bibinfo{journal}{IEEE Geoscience and Remote Sensing Letters}
  \bibinfo{volume}{18}, \bibinfo{pages}{1901--1905}.
\newblock \DOIprefix\doi{10.1109/LGRS.2020.3009823}.
\bibitem[{Ding et~al.(2020)Ding, Zhang, Shen, Li, Wang, Xu and
  Song}]{ding_accelerating_2020}
\bibinfo{author}{Ding, J.Y.}, \bibinfo{author}{Zhang, C.},
  \bibinfo{author}{Shen, L.}, \bibinfo{author}{Li, S.}, \bibinfo{author}{Wang,
  B.}, \bibinfo{author}{Xu, Y.}, \bibinfo{author}{Song, L.},
  \bibinfo{year}{2020}.
\newblock \bibinfo{title}{Accelerating primal solution findings for mixed
  integer programs based on solution prediction}, in:
  \bibinfo{booktitle}{Proceedings of the AAAI Conference on Artificial
  Intelligence}, pp. \bibinfo{pages}{1452--1459}.
\newblock \bibinfo{note}{Issue: 02}.
\bibitem[{Feng et~al.(2024)Feng, Li and Xu}]{FENG2024102362}
\bibinfo{author}{Feng, X.}, \bibinfo{author}{Li, Y.}, \bibinfo{author}{Xu, M.},
  \bibinfo{year}{2024}.
\newblock \bibinfo{title}{Multi-satellite cooperative scheduling method for
  large-scale tasks based on hybrid graph neural network and metaheuristic
  algorithm}.
\newblock \bibinfo{journal}{Advanced Engineering Informatics}
  \bibinfo{volume}{60}, \bibinfo{pages}{102362}.
\newblock \DOIprefix\doi{10.1016/j.aei.2024.102362}.
\bibitem[{Filho et~al.(2020)Filho, Nicolau, Paiva and
  Possamai}]{filho_comprehensive_2020}
\bibinfo{author}{Filho, E.M.}, \bibinfo{author}{Nicolau, V.d.P.},
  \bibinfo{author}{Paiva, K.V.d.}, \bibinfo{author}{Possamai, T.S.},
  \bibinfo{year}{2020}.
\newblock \bibinfo{title}{A comprehensive attitude formulation with spin for
  numerical model of irradiance for {CubeSats} and {Picosats}}.
\newblock \bibinfo{journal}{Applied Thermal Engineering} \bibinfo{volume}{168},
  \bibinfo{pages}{114859}.
\newblock \DOIprefix\doi{10.1016/j.applthermaleng.2019.114859}.
\bibitem[{Gasse et~al.(2019)Gasse, Chételat, Ferroni, Charlin and
  Lodi}]{gasse_exact_2019}
\bibinfo{author}{Gasse, M.}, \bibinfo{author}{Chételat, D.},
  \bibinfo{author}{Ferroni, N.}, \bibinfo{author}{Charlin, L.},
  \bibinfo{author}{Lodi, A.}, \bibinfo{year}{2019}.
\newblock \bibinfo{title}{Exact combinatorial optimization with graph
  convolutional neural networks}.
\newblock \bibinfo{journal}{Advances in Neural Information Processing Systems}
  \bibinfo{volume}{32}.
\bibitem[{Guo et~al.(2023)Guo, Vanhoucke and Coelho}]{guo_prediction_2023}
\bibinfo{author}{Guo, W.}, \bibinfo{author}{Vanhoucke, M.},
  \bibinfo{author}{Coelho, J.}, \bibinfo{year}{2023}.
\newblock \bibinfo{title}{A prediction model for ranking branch-and-bound
  procedures for the resource-constrained project scheduling problem}.
\newblock \bibinfo{journal}{European Journal of Operational Research}
  \bibinfo{volume}{306}, \bibinfo{pages}{579--595}.
\newblock \DOIprefix\doi{10.1016/j.ejor.2022.08.042}.
\bibitem[{Hamilton et~al.(2017)Hamilton, Ying and
  Leskovec}]{hamilton_inductive_2017}
\bibinfo{author}{Hamilton, W.}, \bibinfo{author}{Ying, Z.},
  \bibinfo{author}{Leskovec, J.}, \bibinfo{year}{2017}.
\newblock \bibinfo{title}{Inductive representation learning on large graphs}.
\newblock \DOIprefix\doi{10.48550/arXiv.1706.02216}.
  \bibinfo{note}{arXiv:1706.02216}.
\bibitem[{Han et~al.(2023)Han, Yang, Chen, Zhou, Zhang, Wang, Sun and
  Luo}]{han_gnn-guided_2023}
\bibinfo{author}{Han, Q.}, \bibinfo{author}{Yang, L.}, \bibinfo{author}{Chen,
  Q.}, \bibinfo{author}{Zhou, X.}, \bibinfo{author}{Zhang, D.},
  \bibinfo{author}{Wang, A.}, \bibinfo{author}{Sun, R.}, \bibinfo{author}{Luo,
  X.}, \bibinfo{year}{2023}.
\newblock \bibinfo{title}{A {GNN}-guided predict-and-search framework for
  mixed-integer linear programming}.
\newblock \DOIprefix\doi{10.48550/arXiv.2302.05636}.
  \bibinfo{note}{arXiv:2302.05636}.
\bibitem[{He et~al.(2022)He, Xing, Chen, Pedrycz, Wang and
  Wu}]{heGenericMarkovDecision2022}
\bibinfo{author}{He, Y.}, \bibinfo{author}{Xing, L.}, \bibinfo{author}{Chen,
  Y.}, \bibinfo{author}{Pedrycz, W.}, \bibinfo{author}{Wang, L.},
  \bibinfo{author}{Wu, G.}, \bibinfo{year}{2022}.
\newblock \bibinfo{title}{A generic {Markov} decision process model and
  reinforcement learning method for scheduling agile {Earth} observation
  satellites}.
\newblock \bibinfo{journal}{IEEE Transactions on Systems, Man, and Cybernetics:
  Systems} \bibinfo{volume}{52}, \bibinfo{pages}{1463--1474}.
\newblock \DOIprefix\doi{10.1109/TSMC.2020.3020732}.
\bibitem[{Hopfield and Tank(1985)}]{hopfield_neural_1985}
\bibinfo{author}{Hopfield, J.J.}, \bibinfo{author}{Tank, D.W.},
  \bibinfo{year}{1985}.
\newblock \bibinfo{title}{`{Neural}' computation of decisions in optimization
  problems}.
\newblock \bibinfo{journal}{Biological Cybernetics} \bibinfo{volume}{52},
  \bibinfo{pages}{141--152}.
\newblock \DOIprefix\doi{10.1007/BF00339943}.
\bibitem[{Huang et~al.(2021)Huang, Mu, Wu, Cui and Duan}]{rs13122377}
\bibinfo{author}{Huang, Y.}, \bibinfo{author}{Mu, Z.}, \bibinfo{author}{Wu,
  S.}, \bibinfo{author}{Cui, B.}, \bibinfo{author}{Duan, Y.},
  \bibinfo{year}{2021}.
\newblock \bibinfo{title}{Revising the observation satellite scheduling problem
  based on deep reinforcement learning}.
\newblock \bibinfo{journal}{Remote Sensing} \bibinfo{volume}{13}.
\newblock \DOIprefix\doi{10.3390/rs13122377}.
\bibitem[{Jacquet et~al.(2024)Jacquet, Infantes, Meuleau, Benazera, Roussel,
  Baudoui and Guerra}]{jacquetEarthObservationSatellite2024}
\bibinfo{author}{Jacquet, A.}, \bibinfo{author}{Infantes, G.},
  \bibinfo{author}{Meuleau, N.}, \bibinfo{author}{Benazera, E.},
  \bibinfo{author}{Roussel, S.}, \bibinfo{author}{Baudoui, V.},
  \bibinfo{author}{Guerra, J.}, \bibinfo{year}{2024}.
\newblock \bibinfo{title}{Earth observation satellite scheduling with graph
  neural networks}.
\newblock \DOIprefix\doi{10.48550/arXiv.2408.15041}.
\bibitem[{Jalali Khalil~Abadi et~al.(2024)Jalali Khalil~Abadi, Mansouri and
  Javidi}]{jalalikhalilabadiDeepReinforcementLearningbased2024}
\bibinfo{author}{Jalali Khalil~Abadi, Z.}, \bibinfo{author}{Mansouri, N.},
  \bibinfo{author}{Javidi, M.M.}, \bibinfo{year}{2024}.
\newblock \bibinfo{title}{Deep reinforcement learning-based scheduling in
  distributed systems: A critical review}.
\newblock \bibinfo{journal}{Knowledge and Information Systems}
  \bibinfo{volume}{66}, \bibinfo{pages}{5709--5782}.
\newblock \DOIprefix\doi{10.1007/s10115-024-02167-7}.
\bibitem[{Karimi-Mamaghan et~al.(2022)Karimi-Mamaghan, Mohammadi, Meyer,
  Karimi-Mamaghan and Talbi}]{karimi-mamaghan_machine_2022}
\bibinfo{author}{Karimi-Mamaghan, M.}, \bibinfo{author}{Mohammadi, M.},
  \bibinfo{author}{Meyer, P.}, \bibinfo{author}{Karimi-Mamaghan, A.M.},
  \bibinfo{author}{Talbi, E.G.}, \bibinfo{year}{2022}.
\newblock \bibinfo{title}{Machine learning at the service of meta-heuristics
  for solving combinatorial optimization problems: {A} state-of-the-art}.
\newblock \bibinfo{journal}{European Journal of Operational Research}
  \bibinfo{volume}{296}, \bibinfo{pages}{393--422}.
\newblock \DOIprefix\doi{10.1016/j.ejor.2021.04.032}.
\bibitem[{Kenworthy et~al.(2022)Kenworthy, Nayak, Chin and
  Balakrishnan}]{Kenworthy_Nayak_Chin_Balakrishnan_2022}
\bibinfo{author}{Kenworthy, L.}, \bibinfo{author}{Nayak, S.},
  \bibinfo{author}{Chin, C.}, \bibinfo{author}{Balakrishnan, H.},
  \bibinfo{year}{2022}.
\newblock \bibinfo{title}{{NICE:} {R}obust scheduling through reinforcement
  learning-guided integer programming}.
\newblock \bibinfo{journal}{Proceedings of the AAAI Conference on Artificial
  Intelligence} \bibinfo{volume}{36}, \bibinfo{pages}{9821--9829}.
\newblock \DOIprefix\doi{10.1609/aaai.v36i9.21218}.
\bibitem[{Khalil et~al.(2017)Khalil, Dai, Zhang, Dilkina and
  Song}]{khalil_learning_2017}
\bibinfo{author}{Khalil, E.}, \bibinfo{author}{Dai, H.},
  \bibinfo{author}{Zhang, Y.}, \bibinfo{author}{Dilkina, B.},
  \bibinfo{author}{Song, L.}, \bibinfo{year}{2017}.
\newblock \bibinfo{title}{Learning combinatorial optimization algorithms over
  graphs}, in: \bibinfo{booktitle}{Advances in Neural Information Processing
  Systems}.
\newblock \URLprefix
  \url{https://proceedings.neurips.cc/paper_files/paper/2017/hash/d9896106ca98d3d05b8cbdf4fd8b13a1-Abstract.html}.
\bibitem[{Khalil et~al.(2022)Khalil, Morris and Lodi}]{khalil_mip-gnn_2022}
\bibinfo{author}{Khalil, E.B.}, \bibinfo{author}{Morris, C.},
  \bibinfo{author}{Lodi, A.}, \bibinfo{year}{2022}.
\newblock \bibinfo{title}{{MIP}-{GNN}: {A} data-driven framework for guiding
  combinatorial solvers}.
\newblock \bibinfo{journal}{Proceedings of the AAAI Conference on Artificial
  Intelligence} \bibinfo{volume}{36}, \bibinfo{pages}{10219--10227}.
\newblock \DOIprefix\doi{10.1609/aaai.v36i9.21262}. \bibinfo{note}{number: 9}.
\bibitem[{Kingma and Ba(2015)}]{kingma_adam_2015}
\bibinfo{author}{Kingma, D.P.}, \bibinfo{author}{Ba, J.}, \bibinfo{year}{2015}.
\newblock \bibinfo{title}{Adam: {A} method for stochastic optimization}, in:
  \bibinfo{editor}{Bengio, Y.}, \bibinfo{editor}{LeCun, Y.} (Eds.),
  \bibinfo{booktitle}{3rd International Conference on Learning Representations
  (ICLR)}, \bibinfo{address}{San Diego, CA}.
\newblock \DOIprefix\doi{10.48550/arXiv.1412.6980}.
\bibitem[{Kipf and Welling(2017)}]{kipf_semi-supervised_2017}
\bibinfo{author}{Kipf, T.N.}, \bibinfo{author}{Welling, M.},
  \bibinfo{year}{2017}.
\newblock \bibinfo{title}{Semi-supervised classification with graph
  convolutional networks}.
\newblock \DOIprefix\doi{10.48550/arXiv.1609.02907}.
  \bibinfo{note}{arXiv:1609.02907}.
\bibitem[{Kruber et~al.(2017)Kruber, Lübbecke and
  Parmentier}]{kruber_learning_2017}
\bibinfo{author}{Kruber, M.}, \bibinfo{author}{Lübbecke, M.E.},
  \bibinfo{author}{Parmentier, A.}, \bibinfo{year}{2017}.
\newblock \bibinfo{title}{Learning when to use a decomposition}, in:
  \bibinfo{editor}{Salvagnin, D.}, \bibinfo{editor}{Lombardi, M.} (Eds.),
  \bibinfo{booktitle}{Integration of {AI} and {OR} {Techniques} in {Constraint}
  {Programming}}, \bibinfo{address}{Cham}. pp. \bibinfo{pages}{202--210}.
\newblock \DOIprefix\doi{10.1007/978-3-319-59776-8_16}.
\bibitem[{Li et~al.(2023)Li, Wu, Liao, Fan, Mao and Pedrycz}]{10004750}
\bibinfo{author}{Li, J.}, \bibinfo{author}{Wu, G.}, \bibinfo{author}{Liao, T.},
  \bibinfo{author}{Fan, M.}, \bibinfo{author}{Mao, X.},
  \bibinfo{author}{Pedrycz, W.}, \bibinfo{year}{2023}.
\newblock \bibinfo{title}{Task scheduling under a novel framework for data
  relay satellite network via deep reinforcement learning}.
\newblock \bibinfo{journal}{IEEE Transactions on Vehicular Technology}
  \bibinfo{volume}{72}, \bibinfo{pages}{6654--6668}.
\newblock \DOIprefix\doi{10.1109/TVT.2022.3233358}.
\bibitem[{Lucia et~al.(2021)Lucia, Denby, Manchester, Desai, Ruppel and
  Colin}]{lucia_computational_2021}
\bibinfo{author}{Lucia, B.}, \bibinfo{author}{Denby, B.},
  \bibinfo{author}{Manchester, Z.}, \bibinfo{author}{Desai, H.},
  \bibinfo{author}{Ruppel, E.}, \bibinfo{author}{Colin, A.},
  \bibinfo{year}{2021}.
\newblock \bibinfo{title}{Computational nanosatellite constellations:
  Opportunities and challenges}.
\newblock \bibinfo{journal}{GetMobile: Mobile Comp. and Comm.}
  \bibinfo{volume}{25}, \bibinfo{pages}{16–23}.
\newblock \DOIprefix\doi{10.1145/3471440.3471446}.
\bibitem[{MacFarland and Yates(2016)}]{macfarlandKruskalWallisHTest2016}
\bibinfo{author}{MacFarland, T.W.}, \bibinfo{author}{Yates, J.M.},
  \bibinfo{year}{2016}.
\newblock \bibinfo{title}{Kruskal--{{Wallis H-Test}} for oneway analysis of
  variance ({ANOVA}) by ranks}, in: \bibinfo{editor}{MacFarland, T.W.},
  \bibinfo{editor}{Yates, J.M.} (Eds.), \bibinfo{booktitle}{Introduction to
  Nonparametric Statistics for the Biological Sciences Using R}.
  \bibinfo{publisher}{Springer International Publishing},
  \bibinfo{address}{Cham}, pp. \bibinfo{pages}{177--211}.
\newblock \DOIprefix\doi{10.1007/978-3-319-30634-6_6}.
\bibitem[{de~Magalhães Taveira-Gomes(2017)}]{gomes_reinforcement_2017}
\bibinfo{author}{de~Magalhães Taveira-Gomes, T.S.}, \bibinfo{year}{2017}.
\newblock \bibinfo{title}{Reinforcement Learning for Primary Care and
  Appointment Scheduling}.
\newblock Ph.D. thesis. Faculdade de Engenharia da Universidade do Porto.
\bibitem[{Malitsky et~al.(2016)Malitsky, Merschformann, O’Sullivan and
  Tierney}]{malitsky_structure-preserving_2016}
\bibinfo{author}{Malitsky, Y.}, \bibinfo{author}{Merschformann, M.},
  \bibinfo{author}{O’Sullivan, B.}, \bibinfo{author}{Tierney, K.},
  \bibinfo{year}{2016}.
\newblock \bibinfo{title}{Structure-preserving instance generation}, in:
  \bibinfo{editor}{Festa, P.}, \bibinfo{editor}{Sellmann, M.},
  \bibinfo{editor}{Vanschoren, J.} (Eds.), \bibinfo{booktitle}{Learning and
  {Intelligent} {Optimization}}, \bibinfo{address}{Cham}. pp.
  \bibinfo{pages}{123--140}.
\newblock \DOIprefix\doi{10.1007/978-3-319-50349-3_9}.
\bibitem[{Marcelino et~al.(2020)Marcelino, Vega-Martinez, Seman, Kessler~Slongo
  and Bezerra}]{marcelino_critical_2020}
\bibinfo{author}{Marcelino, G.M.}, \bibinfo{author}{Vega-Martinez, S.},
  \bibinfo{author}{Seman, L.O.}, \bibinfo{author}{Kessler~Slongo, L.},
  \bibinfo{author}{Bezerra, E.A.}, \bibinfo{year}{2020}.
\newblock \bibinfo{title}{A critical embedded system challenge: The
  {FloripaSat}-1 mission}.
\newblock \bibinfo{journal}{IEEE Latin America Transactions}
  \bibinfo{volume}{18}, \bibinfo{pages}{249--256}.
\newblock \DOIprefix\doi{10.1109/TLA.2020.9085277}.
\bibitem[{Nagel et~al.(2020)Nagel, de~Moraes~Novo and Kampel}]{Nagel2020}
\bibinfo{author}{Nagel, G.W.}, \bibinfo{author}{de~Moraes~Novo, E.M.L.},
  \bibinfo{author}{Kampel, M.}, \bibinfo{year}{2020}.
\newblock \bibinfo{title}{Nanosatellites applied to optical {Earth}
  observation: a review}.
\newblock \bibinfo{journal}{Ambiente e Agua -- An Interdisciplinary Journal of
  Applied Science} \bibinfo{volume}{15}, \bibinfo{pages}{1}.
\newblock \DOIprefix\doi{10.4136/ambi-agua.2513}.
\bibitem[{Nair et~al.(2020)Nair, Alizadeh and {others}}]{nair_neural_2020}
\bibinfo{author}{Nair, V.}, \bibinfo{author}{Alizadeh, M.},
  \bibinfo{author}{{others}}, \bibinfo{year}{2020}.
\newblock \bibinfo{title}{Neural large neighborhood search}, in:
  \bibinfo{booktitle}{Learning {Meets} {Combinatorial} {Algorithms} at
  {NeurIPS2020}}.
\bibitem[{Niu et~al.(2018)Niu, Tang and Wu}]{NIU2018813}
\bibinfo{author}{Niu, X.}, \bibinfo{author}{Tang, H.}, \bibinfo{author}{Wu,
  L.}, \bibinfo{year}{2018}.
\newblock \bibinfo{title}{Satellite scheduling of large areal tasks for rapid
  response to natural disaster using a multi-objective genetic algorithm}.
\newblock \bibinfo{journal}{International Journal of Disaster Risk Reduction}
  \bibinfo{volume}{28}, \bibinfo{pages}{813--825}.
\newblock \DOIprefix\doi{10.1016/j.ijdrr.2018.02.013}.
\bibitem[{O'Malley et~al.(2023)O'Malley, {de Mars}, Badesa and
  Strbac}]{omalleyReinforcementLearningMixedinteger2023}
\bibinfo{author}{O'Malley, C.}, \bibinfo{author}{{de Mars}, P.},
  \bibinfo{author}{Badesa, L.}, \bibinfo{author}{Strbac, G.},
  \bibinfo{year}{2023}.
\newblock \bibinfo{title}{Reinforcement learning and mixed-integer programming
  for power plant scheduling in low carbon systems: {{Comparison}} and
  hybridisation}.
\newblock \bibinfo{journal}{Applied Energy} \bibinfo{volume}{349},
  \bibinfo{pages}{121659}.
\newblock \DOIprefix\doi{10.1016/j.apenergy.2023.121659}.
\bibitem[{Pacheco et~al.(2023a)Pacheco, Seman and
  Camponogara}]{pacheco2023deeplearningbased}
\bibinfo{author}{Pacheco, B.M.}, \bibinfo{author}{Seman, L.O.},
  \bibinfo{author}{Camponogara, E.}, \bibinfo{year}{2023}a.
\newblock \bibinfo{title}{Deep-learning-based early fixing for gas-lifted oil
  production optimization: Supervised and weakly-supervised approaches}.
\newblock \href{http://arxiv.org/abs/2309.00197}{{\tt arXiv:2309.00197}}.
\bibitem[{Pacheco et~al.(2023b)Pacheco, Seman, Rigo and
  Camponogara}]{pacheco_bruno_m_2023_8356798}
\bibinfo{author}{Pacheco, B.M.}, \bibinfo{author}{Seman, L.O.},
  \bibinfo{author}{Rigo, C.A.}, \bibinfo{author}{Camponogara, E.},
  \bibinfo{year}{2023}b.
\newblock \bibinfo{title}{[dataset] {FloripaSat MILPs}}.
\newblock \DOIprefix\doi{10.5281/zenodo.8356798}.
\bibitem[{Parmentier and T’Kindt(2023)}]{parmentier_structured_2023}
\bibinfo{author}{Parmentier, A.}, \bibinfo{author}{T’Kindt, V.},
  \bibinfo{year}{2023}.
\newblock \bibinfo{title}{Structured learning based heuristics to solve the
  single machine scheduling problem with release times and sum of completion
  times}.
\newblock \bibinfo{journal}{European Journal of Operational Research}
  \bibinfo{volume}{305}, \bibinfo{pages}{1032--1041}.
\newblock \DOIprefix\doi{10.1016/j.ejor.2022.06.040}.
\bibitem[{Rigo et~al.(2023)Rigo, Morsch~Filho, Seman, Loures and
  Leithardt}]{rigo_instance_2023}
\bibinfo{author}{Rigo, C.A.}, \bibinfo{author}{Morsch~Filho, E.},
  \bibinfo{author}{Seman, L.O.}, \bibinfo{author}{Loures, L.},
  \bibinfo{author}{Leithardt, V.R.Q.}, \bibinfo{year}{2023}.
\newblock \bibinfo{title}{Instance and data generation for the offline
  nanosatellite task scheduling problem}.
\newblock \bibinfo{journal}{Data} \bibinfo{volume}{8}.
\newblock \DOIprefix\doi{10.3390/data8030062}.
\bibitem[{Rigo et~al.(2021a)Rigo, Seman, Camponogara, Filho and
  Bezerra}]{rigo_nanosatellite_2021}
\bibinfo{author}{Rigo, C.A.}, \bibinfo{author}{Seman, L.O.},
  \bibinfo{author}{Camponogara, E.}, \bibinfo{author}{Filho, E.M.},
  \bibinfo{author}{Bezerra, E.A.}, \bibinfo{year}{2021}a.
\newblock \bibinfo{title}{A nanosatellite task scheduling framework to improve
  mission value using fuzzy constraints}.
\newblock \bibinfo{journal}{Expert Systems with Applications}
  \bibinfo{volume}{175}, \bibinfo{pages}{114784}.
\newblock \DOIprefix\doi{10.1016/j.eswa.2021.114784}.
\bibitem[{Rigo et~al.(2021b)Rigo, Seman, Camponogara, Filho and
  Bezerra}]{rigo_task_2021}
\bibinfo{author}{Rigo, C.A.}, \bibinfo{author}{Seman, L.O.},
  \bibinfo{author}{Camponogara, E.}, \bibinfo{author}{Filho, E.M.},
  \bibinfo{author}{Bezerra, E.A.}, \bibinfo{year}{2021}b.
\newblock \bibinfo{title}{Task scheduling for optimal power management and
  quality-of-service assurance in {CubeSats}}.
\newblock \bibinfo{journal}{Acta Astronautica} \bibinfo{volume}{179},
  \bibinfo{pages}{550--560}.
\newblock \DOIprefix\doi{10.1016/j.actaastro.2020.11.016}.
\bibitem[{Rigo et~al.(2022)Rigo, Seman, Camponogara, Filho, Bezerra and
  Munari}]{rigo_branch-and-price_2022}
\bibinfo{author}{Rigo, C.A.}, \bibinfo{author}{Seman, L.O.},
  \bibinfo{author}{Camponogara, E.}, \bibinfo{author}{Filho, E.M.},
  \bibinfo{author}{Bezerra, E.A.}, \bibinfo{author}{Munari, P.},
  \bibinfo{year}{2022}.
\newblock \bibinfo{title}{A branch-and-price algorithm for nanosatellite task
  scheduling to improve mission quality-of-service}.
\newblock \bibinfo{journal}{European Journal of Operational Research}
  \bibinfo{volume}{303}, \bibinfo{pages}{168--183}.
\newblock \DOIprefix\doi{10.1016/j.ejor.2022.02.040}.
\bibitem[{Rouzot et~al.(2024)Rouzot, Gobert, Artigues, Boyer, Camps, Garnier,
  Hebrard and Lopez}]{rouzot:hal-04430171}
\bibinfo{author}{Rouzot, J.}, \bibinfo{author}{Gobert, J.},
  \bibinfo{author}{Artigues, C.}, \bibinfo{author}{Boyer, R.},
  \bibinfo{author}{Camps, F.}, \bibinfo{author}{Garnier, P.},
  \bibinfo{author}{Hebrard, E.}, \bibinfo{author}{Lopez, P.},
  \bibinfo{year}{2024}.
\newblock \bibinfo{title}{{Scheduling onboard tasks of the NIMPH
  nanosatellite}}, in: \bibinfo{editor}{Library, S.D.} (Ed.),
  \bibinfo{booktitle}{{13th International Conference on Operations Research and
  Enterprise Systems (ICORES)}}, \bibinfo{address}{Rome, Italy}.
\newblock \URLprefix \url{https://laas.hal.science/hal-04430171}.
\bibitem[{Saeed et~al.(2020)Saeed, Elzanaty, Almorad, Dahrouj, Al-Naffouri and
  Alouini}]{saeed_cubesat_2020}
\bibinfo{author}{Saeed, N.}, \bibinfo{author}{Elzanaty, A.},
  \bibinfo{author}{Almorad, H.}, \bibinfo{author}{Dahrouj, H.},
  \bibinfo{author}{Al-Naffouri, T.Y.}, \bibinfo{author}{Alouini, M.S.},
  \bibinfo{year}{2020}.
\newblock \bibinfo{title}{Cubesat communications: Recent advances and future
  challenges}.
\newblock \bibinfo{journal}{IEEE Communications Surveys \& Tutorials}
  \bibinfo{volume}{22}, \bibinfo{pages}{1839--1862}.
\newblock \DOIprefix\doi{10.1109/COMST.2020.2990499}.
\bibitem[{Seman et~al.(2022)Seman, Ribeiro, Rigo, Filho, Camponogara, Leonardi
  and Bezerra}]{seman_energy-aware_2022}
\bibinfo{author}{Seman, L.O.}, \bibinfo{author}{Ribeiro, B.F.},
  \bibinfo{author}{Rigo, C.A.}, \bibinfo{author}{Filho, E.M.},
  \bibinfo{author}{Camponogara, E.}, \bibinfo{author}{Leonardi, R.},
  \bibinfo{author}{Bezerra, E.A.}, \bibinfo{year}{2022}.
\newblock \bibinfo{title}{An energy-aware task scheduling for
  quality-of-service assurance in constellations of nanosatellites}.
\newblock \bibinfo{journal}{Sensors} \bibinfo{volume}{22}.
\newblock \DOIprefix\doi{10.3390/s22103715}.
\bibitem[{Seman et~al.(2023)Seman, Rigo, Camponogara, Bezerra and dos
  Santos~Coelho}]{SEMAN2023110475}
\bibinfo{author}{Seman, L.O.}, \bibinfo{author}{Rigo, C.A.},
  \bibinfo{author}{Camponogara, E.}, \bibinfo{author}{Bezerra, E.A.},
  \bibinfo{author}{dos Santos~Coelho, L.}, \bibinfo{year}{2023}.
\newblock \bibinfo{title}{Explainable column-generation-based genetic algorithm
  for knapsack-like energy aware nanosatellite task scheduling}.
\newblock \bibinfo{journal}{Applied Soft Computing} \bibinfo{volume}{144},
  \bibinfo{pages}{110475}.
\newblock \DOIprefix\doi{10.1016/j.asoc.2023.110475}.
\bibitem[{Shiroma et~al.(2011)Shiroma, Martin, Akagi, Akagi, Wolfe, Fewell and
  Ohta}]{shiroma_cubesats_2011}
\bibinfo{author}{Shiroma, W.A.}, \bibinfo{author}{Martin, L.K.},
  \bibinfo{author}{Akagi, J.M.}, \bibinfo{author}{Akagi, J.T.},
  \bibinfo{author}{Wolfe, B.L.}, \bibinfo{author}{Fewell, B.A.},
  \bibinfo{author}{Ohta, A.T.}, \bibinfo{year}{2011}.
\newblock \bibinfo{title}{{CubeSats}: {A} bright future for nanosatellites}.
\newblock \bibinfo{journal}{Central European Journal of Engineering}
  \bibinfo{volume}{1}, \bibinfo{pages}{9--15}.
\newblock \DOIprefix\doi{10.2478/s13531-011-0007-8}.
\bibitem[{Smith(1999)}]{smith_neural_1999}
\bibinfo{author}{Smith, K.A.}, \bibinfo{year}{1999}.
\newblock \bibinfo{title}{Neural networks for combinatorial optimization: A
  review of more than a decade of research}.
\newblock \bibinfo{journal}{INFORMS Journal on Computing} \bibinfo{volume}{11},
  \bibinfo{pages}{15--34}.
\newblock \DOIprefix\doi{10.1287/ijoc.11.1.15}.
\bibitem[{Smith-Miles and Bowly(2015)}]{smith-miles_generating_2015}
\bibinfo{author}{Smith-Miles, K.}, \bibinfo{author}{Bowly, S.},
  \bibinfo{year}{2015}.
\newblock \bibinfo{title}{Generating new test instances by evolving in instance
  space}.
\newblock \bibinfo{journal}{Computers \& Operations Research}
  \bibinfo{volume}{63}, \bibinfo{pages}{102--113}.
\newblock \DOIprefix\doi{10.1016/j.cor.2015.04.022}.
\bibitem[{Tassel et~al.(2021)Tassel, Gebser and
  Schekotihin}]{tassel2021reinforcement}
\bibinfo{author}{Tassel, P.}, \bibinfo{author}{Gebser, M.},
  \bibinfo{author}{Schekotihin, K.}, \bibinfo{year}{2021}.
\newblock \bibinfo{title}{A reinforcement learning environment for job-shop
  scheduling}.
\newblock \href{http://arxiv.org/abs/2104.03760}{{\tt arXiv:2104.03760}}.
\bibitem[{Wang et~al.(2016)Wang, Demeulemeester and Qiu}]{wang_pure_2016}
\bibinfo{author}{Wang, J.}, \bibinfo{author}{Demeulemeester, E.},
  \bibinfo{author}{Qiu, D.}, \bibinfo{year}{2016}.
\newblock \bibinfo{title}{A pure proactive scheduling algorithm for multiple
  earth observation satellites under uncertainties of clouds}.
\newblock \bibinfo{journal}{Computers \& Operations Research}
  \bibinfo{volume}{74}, \bibinfo{pages}{1--13}.
\newblock \DOIprefix\doi{10.1016/j.cor.2016.04.014}.
\bibitem[{Wang et~al.(2013)Wang, Zhu, Qiu and Yang}]{wang_dynamic_2013}
\bibinfo{author}{Wang, J.}, \bibinfo{author}{Zhu, X.}, \bibinfo{author}{Qiu,
  D.}, \bibinfo{author}{Yang, L.T.}, \bibinfo{year}{2013}.
\newblock \bibinfo{title}{Dynamic scheduling for emergency tasks on distributed
  imaging satellites with task merging}.
\newblock \bibinfo{journal}{IEEE Transactions on Parallel and Distributed
  Systems} \bibinfo{volume}{25}, \bibinfo{pages}{2275--2285}.
\newblock \DOIprefix\doi{10.1109/TPDS.2013.156}.
\bibitem[{Wang et~al.(2021)Wang, Pan and Wang}]{wang_complex_2021}
\bibinfo{author}{Wang, L.}, \bibinfo{author}{Pan, Z.}, \bibinfo{author}{Wang,
  J.}, \bibinfo{year}{2021}.
\newblock \bibinfo{title}{A review of reinforcement learning based intelligent
  optimization for manufacturing scheduling}.
\newblock \bibinfo{journal}{Complex System Modeling and Simulation}
  \bibinfo{volume}{1}, \bibinfo{pages}{257--270}.
\newblock \DOIprefix\doi{10.23919/CSMS.2021.0027}.
\bibitem[{{Waubert de Puiseau} et~al.(2022){Waubert de Puiseau}, Meyes and
  Meisen}]{waubertdepuiseauReliabilityReinforcementLearning2022}
\bibinfo{author}{{Waubert de Puiseau}, C.}, \bibinfo{author}{Meyes, R.},
  \bibinfo{author}{Meisen, T.}, \bibinfo{year}{2022}.
\newblock \bibinfo{title}{On reliability of reinforcement learning based
  production scheduling systems: A comparative survey}.
\newblock \bibinfo{journal}{Journal of Intelligent Manufacturing}
  \bibinfo{volume}{33}, \bibinfo{pages}{911--927}.
\newblock \DOIprefix\doi{10.1007/s10845-022-01915-2}.
\bibitem[{Wilcoxon(1945)}]{wilcoxon_1945}
\bibinfo{author}{Wilcoxon, F.}, \bibinfo{year}{1945}.
\newblock \bibinfo{title}{Individual comparisons by ranking methods}.
\newblock \bibinfo{journal}{Biometrics Bulletin} \bibinfo{volume}{1},
  \bibinfo{pages}{80--83}.
\newblock \URLprefix \url{http://www.jstor.org/stable/3001968}.
\bibitem[{Yang et~al.(2022)Yang, Boland, Dilkina and
  Savelsbergh}]{yang_learning_2022}
\bibinfo{author}{Yang, Y.}, \bibinfo{author}{Boland, N.},
  \bibinfo{author}{Dilkina, B.}, \bibinfo{author}{Savelsbergh, M.},
  \bibinfo{year}{2022}.
\newblock \bibinfo{title}{Learning generalized strong branching for set
  covering, set packing, and 0–1 knapsack problems}.
\newblock \bibinfo{journal}{European Journal of Operational Research}
  \bibinfo{volume}{301}, \bibinfo{pages}{828--840}.
\newblock \DOIprefix\doi{10.1016/j.ejor.2021.11.050}.
\bibitem[{Yehuda et~al.(2020)Yehuda, Gabel and Schuster}]{yehuda_its_2020}
\bibinfo{author}{Yehuda, G.}, \bibinfo{author}{Gabel, M.},
  \bibinfo{author}{Schuster, A.}, \bibinfo{year}{2020}.
\newblock \bibinfo{title}{It’s not what machines can learn, it’s what we
  cannot teach}, in: \bibinfo{booktitle}{Proceedings of the 37th
  {International} {Conference} on {Machine} {Learning}},
  \bibinfo{publisher}{PMLR}. pp. \bibinfo{pages}{10831--10841}.
\newblock \URLprefix \url{https://proceedings.mlr.press/v119/yehuda20a.html}.
  \bibinfo{note}{iSSN: 2640-3498}.
\bibitem[{Zhang et~al.(2020)Zhang, Song, Cao, Zhang, Tan and
  Chi}]{NEURIPS2020_11958dfe}
\bibinfo{author}{Zhang, C.}, \bibinfo{author}{Song, W.}, \bibinfo{author}{Cao,
  Z.}, \bibinfo{author}{Zhang, J.}, \bibinfo{author}{Tan, P.S.},
  \bibinfo{author}{Chi, X.}, \bibinfo{year}{2020}.
\newblock \bibinfo{title}{Learning to dispatch for job shop scheduling via deep
  reinforcement learning}, in: \bibinfo{editor}{Larochelle, H.},
  \bibinfo{editor}{Ranzato, M.}, \bibinfo{editor}{Hadsell, R.},
  \bibinfo{editor}{Balcan, M.}, \bibinfo{editor}{Lin, H.} (Eds.),
  \bibinfo{booktitle}{Advances in Neural Information Processing Systems},
  \bibinfo{publisher}{Curran Associates, Inc.}. pp.
  \bibinfo{pages}{1621--1632}.
\newblock \URLprefix
  \url{https://proceedings.neurips.cc/paper_files/paper/2020/file/11958dfee29b6709f48a9ba0387a2431-Paper.pdf}.
\bibitem[{Zhang et~al.(2023)Zhang, Liu, Li, Zhen, Yuan, Li and
  Yan}]{zhang_survey_2023}
\bibinfo{author}{Zhang, J.}, \bibinfo{author}{Liu, C.}, \bibinfo{author}{Li,
  X.}, \bibinfo{author}{Zhen, H.L.}, \bibinfo{author}{Yuan, M.},
  \bibinfo{author}{Li, Y.}, \bibinfo{author}{Yan, J.}, \bibinfo{year}{2023}.
\newblock \bibinfo{title}{A survey for solving mixed integer programming via
  machine learning}.
\newblock \bibinfo{journal}{Neurocomputing} \bibinfo{volume}{519},
  \bibinfo{pages}{205--217}.
\newblock \DOIprefix\doi{10.1016/j.neucom.2022.11.024}.

\end{thebibliography}






\end{document}